\documentclass[final,5p,times,twocolumn]{elsarticle}

\usepackage{graphicx}
\usepackage{amssymb}
\usepackage{caption}
\usepackage{subcaption}
\usepackage{rotating}
\usepackage{float}
\usepackage{acronym}
\usepackage{booktabs}
\usepackage{comment}
\usepackage{titlesec}

\usepackage[section]{placeins}

\titleformat{\paragraph}
{\normalfont\normalsize\itshape}{\theparagraph}{1em}{}
\titlespacing*{\paragraph}
{0pt}{3.25ex plus 1ex minus .2ex}{1.5ex plus .2ex}

\usepackage{amsmath}
\usepackage[super]{nth}
\usepackage{longtable}
\usepackage{tocloft}

\usepackage{array}
\newcolumntype{L}{>{\centering\arraybackslash}m{3cm}}

\usepackage[colorlinks=true]{hyperref}

\usepackage{stmaryrd}

\usepackage{latexsym}
\usepackage{pifont}
\usepackage{xcolor}
\newcommand{\cmark}{\textcolor{olive}{\ding{51}}}%
\newcommand{\xmark}{\textcolor{red}{\ding{55}}}%
\usepackage{multirow}
\usepackage{subcaption}
\usepackage{enumerate}   
\usepackage{courier}
\usepackage[section]{placeins}

\biboptions{authoryear}

\journal{to Remote Sensing of Environment}

\acrodef{AUC}{Area Under the Curve }
\acrodef{AWS}{Amazon Web Services}
\acrodef{ANN}{Artificial Neural Network}
\acrodef{BDAP}{“Big Data Analytics Platform”}
\acrodef{biLSTM}{bidirectional LSTM}
\acrodef{BoC}{Bag-of-Crops}
\acrodef{BRDF}{Bidirectional Reflection Distribution Function}
\acrodef{BV}{Biophysical variables}
\acrodef{BDAP}{Data Analytics Platform}
\acrodef{C}{Crop Rotations as embeddings}
\acrodef{CAP}{Common Agricultural Policy}
\acrodef{CD}{Crop Distribution}
\acrodef{CDL}{Cropland Data Layer}
\acrodef{CR}{Crop Rotation}
\acrodef{CNN}{Convolutional Neural Networks}
\acrodef{CNN-LSTM}{Convolutional Neural Network-Long-Short-Term-Memory}
\acrodef{EO}{Earth Observation}
\acrodef{EU}{European Union}
\acrodef{F1}{F1 Score}
\acrodef{FAPAR}{Fraction of Absorbed Photosynthetically Active Radiation}
\acrodef{FPR}{False Positive Rate}
\acrodef{FOI}{Feature Of Interest}
\acrodef{FR}{France}
\acrodef{GEO}{Group on Earth Observation}
\acrodef{GSA}{Geospatial Aid Application}
\acrodef{HLS}{Harmonized Landsat Sentinel-2}
\acrodef{IntraYE}{Intra-Year Encoder}
\acrodef{InterYE}{Inter-Year Encoder}
\acrodef{JEODPP}{Joint Research Center Earth Observation Data and Processing Platform}
\acrodef{JECAM}{Joint Experiment of Crop Assessment and Monitoring}
\acrodef{JRC}{Joint Research Centre}
\acrodef{L8}{Landsat 8}
\acrodef{LSTM}{Long-Short-Term-Memory}
\acrodef{LAI}{Leaf Area Index}
\acrodef{LPIS}{Land Parcel Identification System}
\acrodef{M-F1}{Macro F1 Score}
\acrodef{m-F1}{micro-F1 score}
\acrodef{NL}{Netherlands}
\acrodef{NIR}{Near Infrared bands}
\acrodef{PA}{Producer Accuracy}
\acrodef{PSE-LTAE}{Pixel-Set Encoder with a Lightweight Temporal Attention Encoder}
\acrodef{PA}{Producer Accuracy}
\acrodef{QA}{Quality Assessment }
\acrodef{UA}{User Accuracy}
\acrodef{RS}{Remote Sensing}
\acrodef{RNN}{Recurrent Neural Network}
\acrodef{RNN-LSTM}{Recurrent Neural Network-Long-Short-Term-Memory}
\acrodef{S1}{Sentinel-1}
\acrodef{S2}{Sentinel-2}
\acrodef{SITS}{Satellite Image Time Series}

\begin{document}

\begin{frontmatter}

\title{Boosting Crop Classification by Hierarchically Fusing Satellite, Rotational, and Contextual Data}


\author{Valentin Barriere$^{1,2,*}$ , Martin Claverie{$^{2,*}$}, Maja Schneider$^{3}$,\\
Guido Lemoine$^{2}$, 
Rapha\"{e}l d'Andrimont\texorpdfstring{$^{2}$}}
\address{ 
$^{1}$\quad Centro Nacional de Inteligencia Artificial (CENIA), Santiago , Chile \\
$^{2}$\quad European Commission, Joint Research Centre (JRC), Ispra , Italy \\
$^{3}$\quad Technical University of Munich, Munich, Germany \\

 $^{*}$ Shared first authorship.
}

\begin{abstract}
Accurate in-season crop type classification is crucial for the crop production estimation and monitoring of agricultural parcels. However, the complexity of the plant growth patterns and their spatio-temporal variability present significant challenges.
While current deep learning-based methods show promise in crop type classification from single- and multi-modal time series, most existing methods rely on a single modality, such as satellite optical remote sensing data or crop rotation patterns. We propose a novel approach to fuse multimodal information into a model for improved accuracy and robustness across multiple crop seasons and countries. 
The approach relies on three modalities used: remote sensing time series from Sentinel-2 and Landsat 8 observations, parcel crop rotation and local crop distribution.
To evaluate our approach, we release a new annotated dataset of 7.4 million agricultural parcels in France (FR) and Netherlands (NL). We associate each parcel with time-series of surface reflectance (Red and NIR) and biophysical variables (LAI, FAPAR). Additionally, we propose a new approach to automatically aggregate crop types into a hierarchical class structure for meaningful model evaluation and a novel data-augmentation technique for early-season classification.
Performance of the multimodal approach was assessed at different aggregation levels in the semantic domain, yielding to various ranges of number of classes spanning from 151 to 8 crop types or groups. It resulted in accuracy ranging from 91\% to 95\% for NL dataset and from 85\% to 89\% for FR dataset.  
Pre-training on a dataset improves domain adaptation between countries, allowing for cross- domain and label prediction, and robustness of the performances in a few-shot setting from France to Netherlands i.e. when the domain changes as per with significantly new labels . 
Our proposed approach outperforms comparable methods by enabling learning methods to use the often overlooked spatio-temporal context of parcels, resulting in increased precision and generalization capacity.
\end{abstract}

\begin{keyword}
Agriculture  
\sep Crop type mapping
\sep Earth Observation 
\sep Geospatial Application 
\sep Long-Short-Term-Memory 
\sep Sentinel-2 
\end{keyword}

\end{frontmatter}





\renewcommand{\thetable}{Table \arabic{table}}
\renewcommand{\thefigure}{Fig. \arabic{figure}}
\renewcommand{\figurename}{}
\renewcommand{\tablename}{}

\newpage
\section{Introduction}

Crop-type maps are an essential element used in crop production monitoring that feed into global food security assessments \citep{porter2014food}. 
Satellite \ac{EO} systems offer a valuable data source for crop classification due to the synoptic, repeated, consistent, and timely availability of observations \citep{weiss2020remote}. Since 2015, the data from the \ac{EU}'s Copernicus program, in particular those of the \ac{S1} and \ac{S2} sensors, provide systematic and consistent \ac{EO} data at a spatial resolution generally higher than the size of most agricultural parcels. Benefiting from satellite image time series (SITS) organised in datasets that were developed over the last decade \citep[see][for a review of existing datasets]{Capliez2023}, many crop-type mapping studies and operational systems based on \ac{EO} have been carried out, leveraging abundant available data. %

Most of the current systems from crop classification rely on the \ac{RS} data of the season (Section \ref{subsubsec:eo-based-sota}).
What we propose in this work is the conjunct use of \ac{CR} with \ac{RS} data, and also local \ac{CD}. If this has been done before, like we will show in the Section \ref{subsubsec:cr-based-sota}, none has tried to fuse these different modalities in an elegant way. The current approaches are mainly focusing on measuring the intra-year dynamics, for example by only inferring using the \ac{RS} data of the target season, while forgoing the inter-year dynamics that is crucial since reflecting agricultural practices. 
By fusing the different data sources in a hierarchical way, the approach described in this paper allows one to take advantage of both the inter-year dynamics and the intra-year dynamics. For this aim, we construct and release a dataset of several seasons of time-series of the same parcels, based on data from the \ac{EU}'s Common Agricultural Policy. 
Finally, and because we judge this important for the practitioners, we propose a method for early-season detection works well with the proposed hierarchical model (Section \ref{subsubsec:es-sota}). 
The chosen encoders are arbitrary, and our method could work with other more complex methods than the \ac{RNN} - \ac{LSTM} (i.e., RNN-LSTM) we used in this work. 

\subsection{Related Works}
The related works have been separated in three subsections: EO-based-only models, models using crop-rotations, and works on early-season predictions. We restrain this section to learning methods, as it is the focus of this work and as they have proven to lead to better results than classical machine learning and signal processing-based methods at large scale.

\subsubsection{EO-based Models} \label{subsubsec:eo-based-sota}

\cite{Russwurm2019} classify 13 crop types at the parcel-level using all 13 available spectral bands of Sentinel2 during the 2017 growing season in French Brittany. They compare a Transformer-Encoder \citep{Vaswani2017} and a Recurrent Neural Network of \ac{LSTM} type \citep{Hochreiter1997} and find that both models perform similarly, with the Transformer-Encoder and \ac{LSTM} achieving both comparable  accuracy and macro-F1, respectively close to 0.69 and 0.59.
\cite{Russwurm2020} design a crop classifier at the parcel-level using \ac{S2} data from three regions of Germany and compare different approaches to model the signal, including a Transformer and an \ac{LSTM}. They achieve overall accuracies between 0.85 and 0.92 using the \ac{LSTM}, depending on the number of classes considered. They conclude that data processing was useful for those kind of models. A similar approach was taken by \cite{Russwurm2019b} on 40k parcels in Central Europe using \ac{S2} data, for which they proposed a new early classification mechanism to enhance a classical model with an additional stopping probability based on previously seen information.
Furthermore, \cite{Russwurm2018} 
tackled the task of crop classification at the pixel level, by accounting for the spatial variation to detect parcels boundaries, using \ac{CNN-LSTM} on \ac{S2} images to classify 17 types of crops in a unique German region.  
\cite{SainteFareGarnot2019} proposed to use a \ac{CNN} before a \ac{RNN}  to learn the aggregation of the parcel pixels instead of classically averaging them, and applied their system on 200k parcels of the south-west of \ac{FR}.
In the end, \cite{SainteFareGarnot2020} proposed a smart method to tackle parcel-level crop classification, by randomly sampling pixels of the parcels to learn expressive descriptors that are processed by a transformer. The application of their models was carried out on 191k parcels located in the south of \ac{FR}, encompassing 20 crop classes.

Finally, only some works attempt few-shot classification (i.e., learning a classifier for a new dataset given only a few examples, zero-shot when no examples are available \citep{Peng2018}) with EO-data because it has been a difficult task for a long-time, knowing that a majority of the systems work poorly without domain data. Nevertheless, \cite{Russwurm2019meta} and \cite{Tseng2021a} both propose to use the MAML meta-learning algorithm in order to tackle few-shot crop or land cover classification at the pixel-level, using \ac{EO} data only. The former on the Sen12MS \citep{Schmitt2019} and DeepGlobe Challenge \citep{Demir2018} datasets for land cover classification. The latter on the \textit{CropHarvest} dataset from \cite{Tseng2021}, which is an aggregation of satellite datasets for crop type classification containing annotation at the pixel levels, without an harmonized label taxonomy between the examples of different domains.

\subsubsection{Crop-rotation-based Models} \label{subsubsec:cr-based-sota}
Crop rotation is an essential agronomic practice for sustainable farming and preserving long-term soil quality. A good understanding and design of crop rotation is vital for sustainability and mitigating the variability of agricultural productivity induced by climate change \citep{BOHAN2021169}. Crop rotation patterns are complex and non-stable in time, often dependent on farmer management decisions and subject to changes due to economic considerations and administrative regulations \citep{dogliotti2003rotat}. As a result, expert knowledge-based models have limitations in terms of accuracy and applicability over large areas and long periods. Alternative approaches, such as estimation of crop sequence probabilities using survey data and hidden Markov models have been demonstrated in \ac{FR} \citep{xiao2014modeling}, but these methods are not always feasible at large scale due to the extended size of the required sample.

Past research has focused on using machine learning techniques to predict crop rotations. In \cite{Osman2015}, a Markov Logic model is used to predict the following season's crop in \ac{FR}, achieving an accuracy of 60\%. Other studies have utilized deep neural networks, such as \cite{Yaramasu2020}, which reaches a maximum accuracy of 88\% on a 6-class portion of the US \ac{CDL} dataset over 12 years.

Only three studies \citep{johnson2021pre,Giordano2020,Quinton2021} have been identified that combine the use of crop rotations and satellite time-series data with deep learning. \cite{johnson2021pre} applied this method over several seasons to derive near real-time \ac{CDL}. However, this methodology is constrained to a small number of crop types and the use of a Random Forest classifier, while recent advancements in deep learning have shown significant improvements in such high-data regime problems. \cite{Giordano2020} used Conditional Random Fields to model the temporal dynamics of crop rotations. They focused on two French regions with very different climate conditions and agricultural practices, using around 9,230 and 1,902 parcels with 2 seasons of data.  
\cite{Quinton2021} propose to use a \ac{PSE-LTAE} \citep{SainteFareGarnot2020} combined with a multi-year classification method. They represent the past crops with a one-hot encoder that they sum, without modeling the dynamics of the sequence. 
In our work, we not only focus on modeling the sequential aspects of crop rotations, but also incorporate the \ac{RS} signals from previous seasons. 

\subsubsection{Early Season Classification} \label{subsubsec:es-sota}
While end-of-season crop type maps have a great interest for agricultural land monitoring \citep{weiss2020remote}, in-season crop production monitoring requires a more rapid response, including before-harvest crop type map releases. 
Some works have also focused on tackling early-season classification. \cite{Russwurm2019b} proposed to solve the problem in an elegant way, with an adapted cost function that only rewards the classifier for an early classification if the right class has been predicted with a respectable degree of accuracy. They extend this work in \cite{Russwurm2023} by presenting end-to-end Learned Early Classification of Time Series, also classifying crops at the parcel-level in France, Germany, Ghana and South Sudan. 
Finally, \cite{lin2022early} proposed an original topology-based approach to automatically label instances of the test season in early-season. They focus on maximum or less than 4 classes at the pixel-level, training a random forest with the test season pseudo-labels obtained previously. 

Without using a special cost function, \cite{Weilandt2023} use a \ac{PSE-LTAE} with a data-augmentation technique initially proposed by \cite{Barriere2022cdceo} on hierarchical \ac{LSTM} for crop-type classification at the parcel-level. They perform data-augmentation by randomly cropping the end of the \ac{EO} time series during training. The data-augmentation technique boosts the performances of early-season classification. They also compared separate models trained on data cropped up to a unique certain period in the year (i.e. one model for one period), which is not efficient in terms of computation and yielded similar results.

\subsection{Positioning and Objectives} \label{sec:objectives}

To the best of the authors' knowledge, there are some gaps in the existing literature. 

A significant amount of research has focused on using remote sensing to predict crop types at the pixel or parcel level using only \ac{EO} and in-situ observations of the current season, treating the signal as independent from one season to another. 
Other studies have used parcel crop rotations to address preseason prediction of crop types \citep{Osman2015,Yaramasu2020}, but the lack of sufficient information in the signal (i.e. short duration of the time series) limits their performance even when targeting minor classes. 
Although the integration of \ac{RS} data with crop rotations has been investigated in certain studies \citep{Giordano2020,Quinton2021}, no one has yet taken this analysis a step further by incorporating its dynamic modeling.

Over the past five years, a multitude of \ac{SITS} datasets have been made available. In his study, \cite{Selea23} compiles a comprehensive inventory of these datasets, encompassing different regions across Europe. All of these datasets are dependent on the accessibility of \ac{LPIS} and \ac{GSA} public datasets. The dataset we have released to support our development is significantly larger in terms of number of parcel and lenght of the time series than any other existing datasets \citep{Selea23}. The processing of \ac{RS} data, which incorporates a smoothing algorithm, sets this dataset apart from others in its category.
We have collected data for a minimum of 5 seasons and have gathered information on approximately 6.8 million (in \ac{FR}) and 600 thousand (in the \ac{NL}) parcels.  
Because of its diversity, we propose a method to aggregate the crops at the regional-level.

The contributions of this study are 5-fold: 

\begin{enumerate}[(i)]
    \item We release a new dataset of more than 7.4 million parcels with their associated crops, and \ac{RS} signals for the period 2016-2020 in \ac{FR} and \ac{NL}. This allows the integration of crop rotation patterns with the remote sensing signal.  
    \item We create a data- and knowledge-driven technique to automatically group crops together in a meaningful way according to their similarity and importance in the region, for a fairest comparison of classifiers in any crop dataset. It leverages expert knowledge from the EuroCrops \citep{schneider2021eurocrops} taxonomy and derive local crop distribution.
    \item We construct a novel approach for crop type mapping from crop rotations and  \ac{S2} optical time series in a multimodal way using a hierarchical \ac{LSTM} network. This approach is unique in its conception, as it fuses large amounts of temporally fine-grained \ac{EO} data with crop rotation analysis in an advanced deep learning method. The crop rotations and the \ac{S2} time series are enhanced by previous-season crop distributions of the neighbourhing parcels.\footnote{Our method is completely feature-independent and could be used with other bands} 
    \item We develop a data-augmentation technique for the in-season classification, by randomly cropping the end of the \ac{RS} time-series data. This allows our model to classify parcels before the end of the season, a crucial feature for real-life application of crop monitoring.
    \item We assess the cross-domain generalization potential of the framework based on a modified nomenclature of EuroCrops, without using any strategy to mitigate domain gap, target shifts, or handle new classes.
\end{enumerate}

\section{Materials} 

This section presents a description of the study area and the \ac{EO} data processing procedure. 

\subsection{Crop reference data, study area, and harmonization of parcel data}
\label{sec:method_study_area}

The \ac{GSA}  corresponds to the annual crop declarations made by \ac{EU} farmers for \ac{CAP}  area-aid support measures. The electronic \ac{GSA}  records include a spatial delineation of the parcels. A \ac{GSA}  element is always a polygon of an agricultural parcel with one crop (or a single crop group with the same payment eligibility). The \ac{GSA}  is operated at the region or country level in the \ac{EU} 27 member states, resulting in about 60 different designs and implementation schemes over the \ac{EU}. Since these infrastructures are set up in each region, data are not interoperable at the moment, and the legends are not semantically harmonised. Furthermore, only few \ac{EU} member states release \ac{GSA}  data as open data, although the overall trend is towards increasingly opening up the data for public use. 

Some efforts have been made to provide harmonised \ac{GSA}  dataset over the \ac{EU}. AI4boundaries \citep{d2023ai4boundaries} provides harmonized parcel geometries over 7 countries in the \ac{EU} to benchmark method for parcel delineation. EuroCrops \citep{schneider2021eurocrops} proposed a semantic harmonisation framework to harmonise the legend of \ac{GSA}  across different countries. This harmonisation is open source and is maintained by the community \footnote{\url{https://github.com/maja601/EuroCrops}}. While EuroCrops provides a unique effort so far, this work is still in progress especially regarding the time dimension. A recent European Commission Implementing Regulation (EU) 2023/138\footnote{\url{https://eur-lex.europa.eu/eli/reg_impl/2023/138}} identifies a list of specific high-value datasets and the arrangements for their publication. This should be a game changer in the opening of the \ac{GSA}  for public access in the future and thus foster their use for research.


The Hierarchical Crop and Agriculture Taxonomy version 2 (HCATv2) from EuroCrops offers a knowledge graph regrouping crops together in a hierarchical way that is coherent with agricultural practices.
It contains 393 classes, which are defined at six hierarchical levels of which the first two are fixed due to compatibility with other taxonomies. 
For example \texttt{33-01-01-05-01} corresponds to the class \textit{Summer Oats}, which is included in its parent class \texttt{33-01-01-05-00} (\textit{Oats}) and its grand-parent class \texttt{33-01-01-00-00} (\textit{Cereals}).
Nevertheless, it is not possible to compare the labels only using the hierarchy because some branches go to a deeper level than others. For example, the class \textit{Capsicum}
is level-4 and represent 0.004\% of \ac{FR}, which is the same level than the class \textit{Cereal} representing 32\%. 
HCATv2 \citep{schneider2021eurocrops} was used to represent 
a Sankey diagram linking the French GSA (left) and the Dutch GSA (right), using HCATv2 (centre) is represented in \ref{fig:Sankey} using  40 main crop types for each country\footnote{An interactive version of the diagram without class limitation is available on \url{https://jeodpp.jrc.ec.europa.eu/ftp/jrc-opendata/DRLL/CropDeepTrans/data/sankey_All_crops.html}.}.

For this study, \ac{FR} and the \ac{NL} were selected because of i) their open parcel data availability, ii) their \ac{EU} representativeness in covering a latitude range from 40$^{\circ}$  to 55$^{\circ}$  Northern latitude as well as four biogeographic regions (i.e. Oceanic, Continental, Alpine and Mediterranean) and iii) the countries have different size and landscape. Parcel \ac{GSA} data from seasons\footnote{season n means from October n to October n+1} 2015 to 2020 over \ac{FR} and from 2013 to 2020 over the \ac{NL} were collected (\ref{tab:RawGSA}). 

\begin{table}[]
\centering
\caption{Original Geospatial Aid Application (GSA) parcel numbers and area per season used in the study for France (FR) and Netherlands (NL). 
The number of distinct crop types are provided using original GSA (4th column) and harmonized using EuroCrops (5th column). The "stack" lines correspond to the Feature Of Interest (FOI, see section \ref{sec:method_parcel_geomtry}).
}
\label{tab:RawGSA}
\footnotesize
\begin{tabular}{lrrrrrr}
\hline
 Country & Season & RS & \multicolumn{2}{c}{\# distinct crop types} &   Number of  & Total area  \\
  &  &  & original    &  harmonized  & polygons &(1000 ha)  \\
\hline
NL & 2013 & \xmark &  76  & 41  & 762,725  & 1,855  \\
NL & 2014 & \xmark &  75  & 41  & 765,006  & 1,859  \\
NL & 2015 & \xmark &  260 & 117 & 790,930  & 1,873  \\
NL & 2016 & \cmark &  296 & 133 & 786,572  & 1,871  \\
NL & 2017 & \cmark &  300 & 136 & 785,710  & 1,882  \\
NL & 2018 & \cmark &  312 & 135 & 774,822  & 1,871  \\
NL & 2019 & \cmark &  317 & 139 & 772,565  & 1,868  \\
NL & 2020 & \cmark &  326 & 141 & 767,034  & 1,872  \\
\hline
NL & stack & &  401 & 148 & 596,762  & 1,407  \\ 
\hline
FR & 2015 & \xmark &  261 & 150 & 9,434,672 & 27,856 \\
FR & 2016 & \cmark &  261 & 147 & 9,334,043 & 27,876 \\
FR & 2017 & \cmark &  280 & 148 & 9,393,747 & 27,889 \\
FR & 2018 & \cmark &  282 & 149 & 9,517,878 & 27,917 \\
FR & 2019 & \cmark &  241 & 149 & 9,604,463 & 27,960 \\
FR & 2020 & \cmark &  239 & 148 & 9,778,397 & 27,998 \\
\hline
FR & stack & &  319 & 151 & 7,051,683 &25,495 \\ 
\hline
\end{tabular}
\end{table}

\subsection{Geometric minimum common parcel extraction through time}
\label{sec:method_parcel_geomtry}

\begin{figure*}[]
    \centering 
    \includegraphics[width=1.\textwidth]{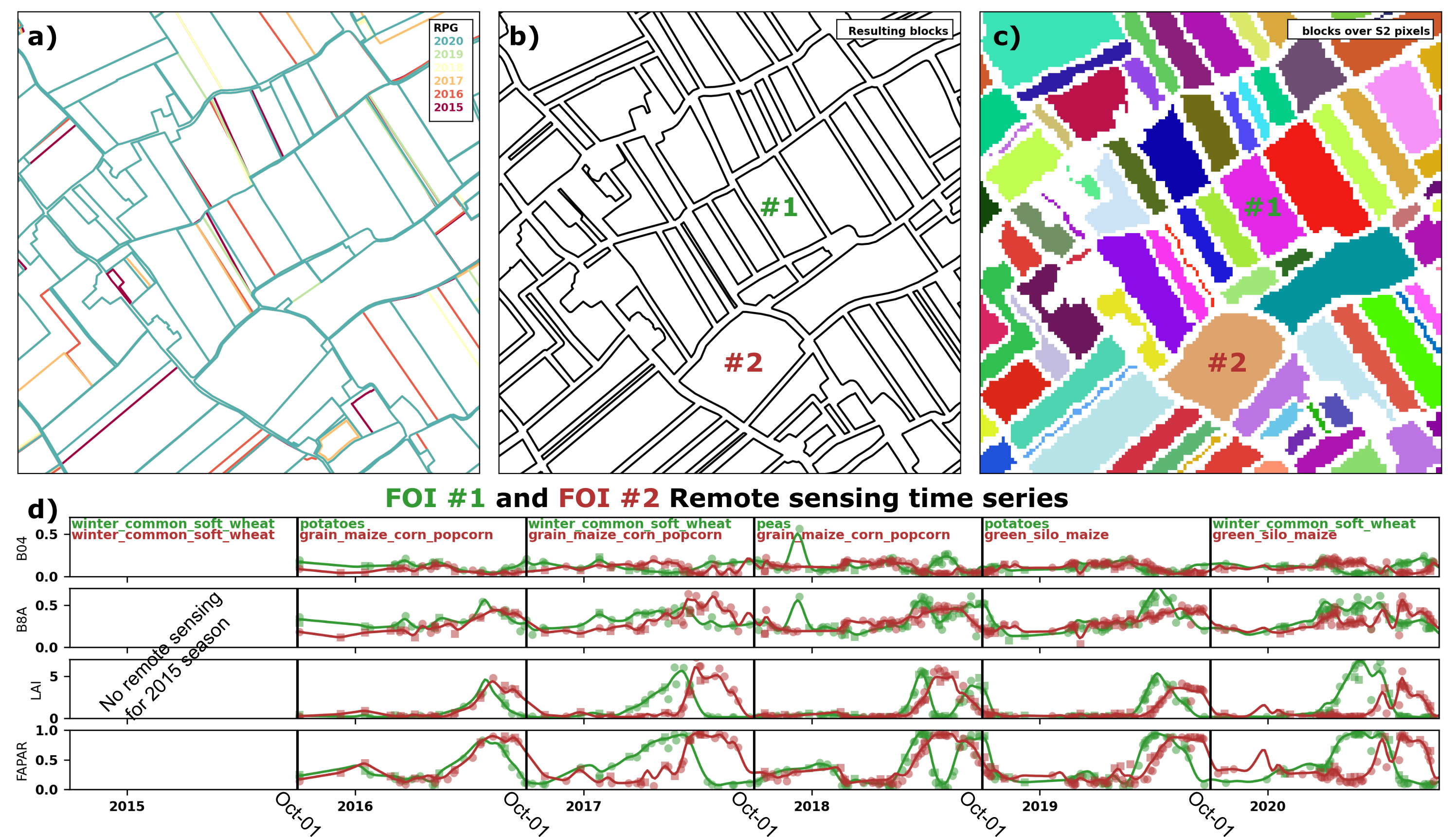} \vspace*{-.5cm}
    \caption{Feature Of Interest (FOI) extraction, time series extraction and smoothing. a) the map shows an overlap of the six \ac{GSA} layers; b) Resulting blocks corresponding to the intersection of the six \ac{GSA} layers, reduced by an inner buffer; c) rasterized version of the blocks used for extracting the \ac{S2} data; d) full \ac{S2} and crop types time series of two selected FOI (shown in panels a-c). Yearly crop types are displayed on the top sub-panel. Input variables time series are displayed using daily observations (circles and squares correspond to \ac{S2} and \ac{L8} data, respectively) and smoothed signal (used as \ac{LSTM} inputs).}
    \label{fig:FOI} \vspace*{-.5cm}
\end{figure*}

\ac{GSA} are delivered yearly as a set of polygons outlining agricultural parcels. From season-to-season, the parcel boundary may change. We intersected GSA data (i.e. 2013-2020 for \ac{NL} and 2015-2020 for FR) in order to extract minimum common area, each with a distinct multi-annual crop sequence, named hereafter \ac{FOI}. Since \ac{FOI} are the cross-section of varying parcel bounds, their overall size is smaller than the annual GSA parcels. We discarded any \ac{FOI} with an area of less than 0.1 ha and 0.5 ha for \ac{NL} and FR, respectively. The total \ac{FOI} area cover 85\% and 93\% of the average GSA area for \ac{NL} and FR respectively (the "stack" entries in \ref{tab:RawGSA}). 
For each \ac{FOI}, a crop type sequence was extracted, as well as the remote sensing time series  (\ref{fig:FOI}). 

\subsection{Earth Observation processing}
\label{sec:method_eo}

Remote sensing data were extracted from \ac{S2} MSI and \ac{L8} OLI sensors. While the GSA data spans from 2013 and 2015, for NL\footnote{It starts from 2009 but we only took data from 2013} and FR respectively, the remote sensing data were used starting 2016 cropping season (i.e., from October-2015), corresponding to the first cropping season with both sensors in-orbit.

\ac{S2} MSI products with Level-2A surface reflectance data were downloaded from the Copernicus Open Access Hub. \ac{L8} OLI surface reflectance data were downloaded from the L30 products of the \ac{HLS} data set. For both products, L2A and L30 \ac{QA} layers were used to mask non-surface-related information. We masked all pixels flagged as cloud, cloud-shadow, cirrus and snow. 

\ac{LAI} and \ac{FAPAR} \ac{BV} maps were derived from the \ac{S2} L2A (20m spatial resolution) and \ac{L8} L30 (30m spatial resolution) products, using the BV-NET algorithm developed by \cite{weiss1999evaluation}. It aims to retrieve the two \ac{BV} from multispectral reflectance using the inversion of the radiative transfer model PROSAIL and a back-propagation \ac{ANN}. Following the configuration of \cite{delloye2018retrieval}, the architecture of the \ac{ANN} consists of two layers: (i) one layer with five tangent sigmoid transfer functions neurons and (ii) one layer with one linear transfer functions neuron. This configuration allows for greater dynamics in the output variables \citep{claverie2013validation}. The \ac{HLS} products are normalised using the \ac{BRDF} with a nadir view zenith angle and a variable sun angle \citep{claverie2018harmonized}, while the \ac{S2} L2A products are unadjusted with \ac{BRDF}. We retained these data specifications and configured two BV-NET models to account for them. For both product types, the cosine of the solar zenith angle was included in the BV-NET input set; for \ac{S2} L2A, the view zenith and relative azimuth angles were also included.

Only the Red and \ac{NIR} were kept for further analysis; the remaining spectral bands were discarded. Four variables (\ac{LAI}, \ac{FAPAR}, Red band and NIR band) pixel-based maps were thus used to derive time series per \ac{FOI}. Pixels whose centres fell within the \ac{FOI} boundaries, reduced by a 15 m inner buffer (to prevent from using mixed pixels and reduce impact of the geometric precision), were averaged using a zonal statistics technique; flagged values (cloud, cloud shadow, cirrus or snow) from \ac{QA} layers were not included in the averaging. \ac{FOI} values were only considered valid if more than 75\% of the \ac{LAI} pixels were valid.

Despite filtering the data using relevant \ac{QA} layers, the resulting \ac{FOI}-based time series are still contaminated by missed cloud, cloud shadow, haze or dense atmosphere. To remove these remaining outliers, we applied a Hampel filter using red and NIR bands to discard cloud and cloud-shadow in the time series respectively; the parameters of the filter follow \cite{claverie2018harmonized}.

Finally, filtered time series of four variables aggregated at \ac{FOI} level were smoothed individually using a Whittaker filter.\footnote{as developed by \url{https://github.com/WFP-VAM/vam.whittaker}} The time series are first gap-filled in time using a linear interpolation and a time step of 2 days. We applied the Whittaker configuration with V-curve optimization of the smoothing parameter and expectile smoothing using asymmetric weight, with an "Envelope" value of 0.9 and a tested lambda range between -1 and 1 \citep{eilers2017automatic}. This results in a smoothed time series with a time step of 4 days and no interruption between the seasons. 

\section{Methods} 

The feature extraction and the model architecture are first described in section \ref{sec:method_lstm_architecture}), followed by a description of the learning model and the integration of features as observations (section \ref{sec:method_hierchical_aggregation}). 
In Section \ref{sec:method_data_augmentation}, we delve into the early-season data augmentation technique. Then in Section \ref{sec:method_transfer_learning}, we explore the training process and the application of models in various countries. Finally, section \ref{sec:method_processing} outlines the processing facilities utilized in the study and includes links to the data and code.

\begin{figure*}[t]
\centering\includegraphics[width=0.8\linewidth]{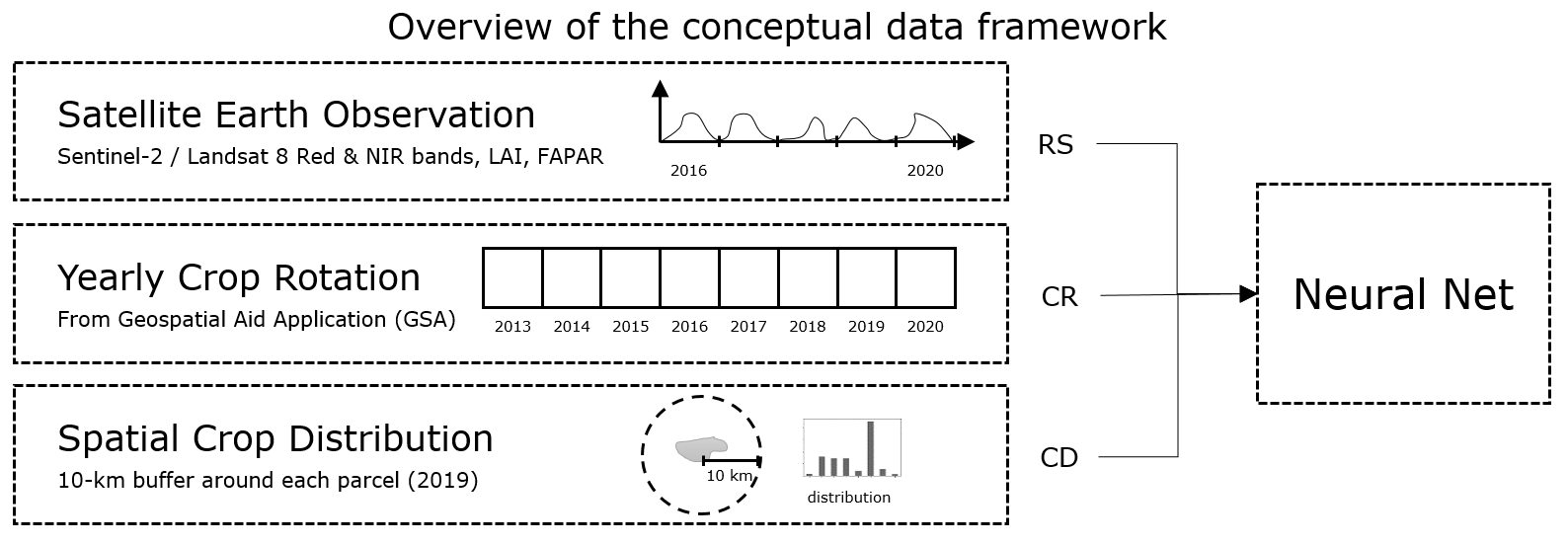}
\caption{Overview of the conceptual data framework  for crop classification to leverage satellite optical time series, yearly crop rotation history, and spatial local crop distributions.}
\label{fig:ConceptualFramework}
\end{figure*}

\subsection{Models description}
\label{sec:method_lstm_architecture}

A series of models were developed involving various configuration and input modalities. The three modalities are \ac{RS}, \ac{CR} and \ac{CD} (see the conceptual framework in \ref{fig:ConceptualFramework}). This section describes the model and the integration of the data as features. 

Crop rotation, defined as sequences of crops throughout the seasons, has been modeled in a manner similar to a sequences of words within a language model \citep{mikolov2010recurrent}. This modeling process is further enhanced by adding \ac{S2} time-series data, which is treated as analogous to the prosody of a speaker \citep{Wollmer2013,wollmer2013youtube,Schuller2016}, i.e. the pattern of intonation, stress and rhythm in a speech.\footnote{The RS encoding could use any other encoder type, like a state-of-the-art model such as PSE-LTAE \citep{SainteFareGarnot2020}} Ultimately, the high-level spatial crop distribution features we add on the last layer of the network can be seen as the distribution of the words generally used by the speaker.

\begin{table*}[]
    \footnotesize
    \caption{Summary of the different models used in this paper, using Crop Rotations (CR), Remote Sensing (RS), and Crop Distribution (CD).} 
    \label{tab:ModelsSummary}
    \centering
    \begin{tabular}{@{}l|ccc|cc|c@{}}
    \toprule
    \multirow{2}{*}{\textbf{Models}}    & \multirow{2}{*}{\textbf{CR}} & \multirow{2}{*}{\textbf{RS}}        & \multirow{2}{*}{\textbf{CD}} & \multicolumn{2}{c|}{\textbf{Modelisation-level}}   & \multirow{2}{*}{\textbf{Hierarchical}}      \\ 
     & & & & \textbf{Within season} & \textbf{Between seasons} & \\ \midrule
    
    $\text{IntraYE}_{RS}$   & \xmark          & \cmark  & \xmark   & \cmark  & \xmark & \xmark \\
    $\text{IntraYE}_{MM}$   & \cmark          & \cmark  & \xmark   & \cmark  & \xmark & \xmark \\ \hline 
    $\text{InterYE}_{Crop}$ & \cmark          & \xmark  & \xmark   & \xmark  & \cmark & \xmark \\
    $\text{InterYE}_{RS}$   & \xmark          & \cmark  & \xmark   & \xmark  & \cmark & \xmark \\
    $\text{InterYE}_{MM}$   & \cmark          & \cmark  & \xmark   & \xmark  & \cmark & \xmark \\ \hline 
    $\text{HierE}_{RS}$     & \xmark          & \cmark  & \xmark   & \cmark  & \cmark & \cmark \\
    $\text{HierE}_{MM}$     & \cmark          & \cmark  & \xmark   & \cmark  & \cmark & \cmark \\
    $\text{HierE}_{final}$  & \cmark          & \cmark  & \cmark   & \cmark  & \cmark & \cmark \\ \bottomrule
    \end{tabular}
\end{table*}

\subsubsection{Features extraction}

\subsubsection*{Crop Rotation}


The crop types labels were extracted from the respective \ac{GSA} and remapped using EuroCrops. This yields to a total of $V_{NL}=141$ and $V_{FR}=151$ classes, for \ac{NL} and FR respectively, corresponding to $V=225$ unique classes. We modeled the crop by a one-hot vector of size $V$ and used it as an input to an embedding layer. For each FOI, we extracted the crop sequence which corresponds to the \ac{CR} feature. 

\subsubsection*{Remote Sensing temporal integration}
We integrated the \ac{RS} time series into features using a sliding window of size $W=1$ month with a step of size $s_w=0.5W$, obtaining a sequence of $t_w=\frac{12}{s_w}=24$ inputs windows for the 12 months of the season, from 1st of October to 31st of September. By utilizing this setup, we obtained some overlap between the windows, which should prevent loss of information by breaking the signal dynamics, albeit with a slight trade-off of redundancy in the features. On every time window, the four RS signals (\ac{LAI}, \ac{FAPAR}, Red band and NIR band) were integrated for each window using seven statistical functionals \citep{Schuller2016} representing the signal as a fixed vector: average mean, standard deviation, min, max, median, first quartile, third quartile. As a total, we obtained $4\times24\times7=672$ features per \ac{FOI} per season. Finally, we normalised each of the 24 features.

\subsubsection*{Crop Distribution}

We computed the total area of each crops in a circle of radius $r=10km$ from each \ac{FOI} and turned it to percentage of the total cropped area. 
The crop type distribution around the \ac{FOI} accounts spatial for variation in terms of agricultural practices in relation with local agro-meteorological conditions, economic and historical factors. In the absence of major shocks, the distribution of the crop types in a region is expected to be stable over the seasons \citep{merlos2020scale}, which determines the \textit{a priori} probability of local crop occurrence. We integrate this local information by adding a vector representing the \ac{CD} over the surrounding crop types centered around each \ac{FOI} centroid. 
The spatial \ac{CD} was always derived from the same season for computational reasons. We used the 2019 validation set season, i.e. not for the test season 2020. We rounded the probability at $10^{-4}$, leading to some values being 0 when not null.
Despite the Eurocrops harmonization, the crop lists of the two datasets (FR and NL) are not identical. We used the union of the two crop lists for both datasets.


\subsubsection{Architecture of the models}
Eight models were developed for the study. Their architectures are presented hereafter and summarized in \ref{tab:ModelsSummary}.

\paragraph*{Baselines using Year-Independent models} \label{subsubsec:other_baselines}
We used two baselines models that treat the \ac{RS} time series signal in a classical way without using hierarchical networks and without modeling the dynamics between the seasons. One unimodal model is using only \ac{RS} data and another one is multimodal using \ac{RS} and \ac{CR}, based respectively on the works of \cite{Russwurm2019} and \cite{Quinton2021}. \textbf{These models are referred to as $\text{IntraYE}_{RS}$ and $\text{IntraYE}_{MM}$, respectively}.

\begin{figure*}[h]
    \centering 
    \hspace*{-1cm}
    \includegraphics[width=1.1\textwidth]{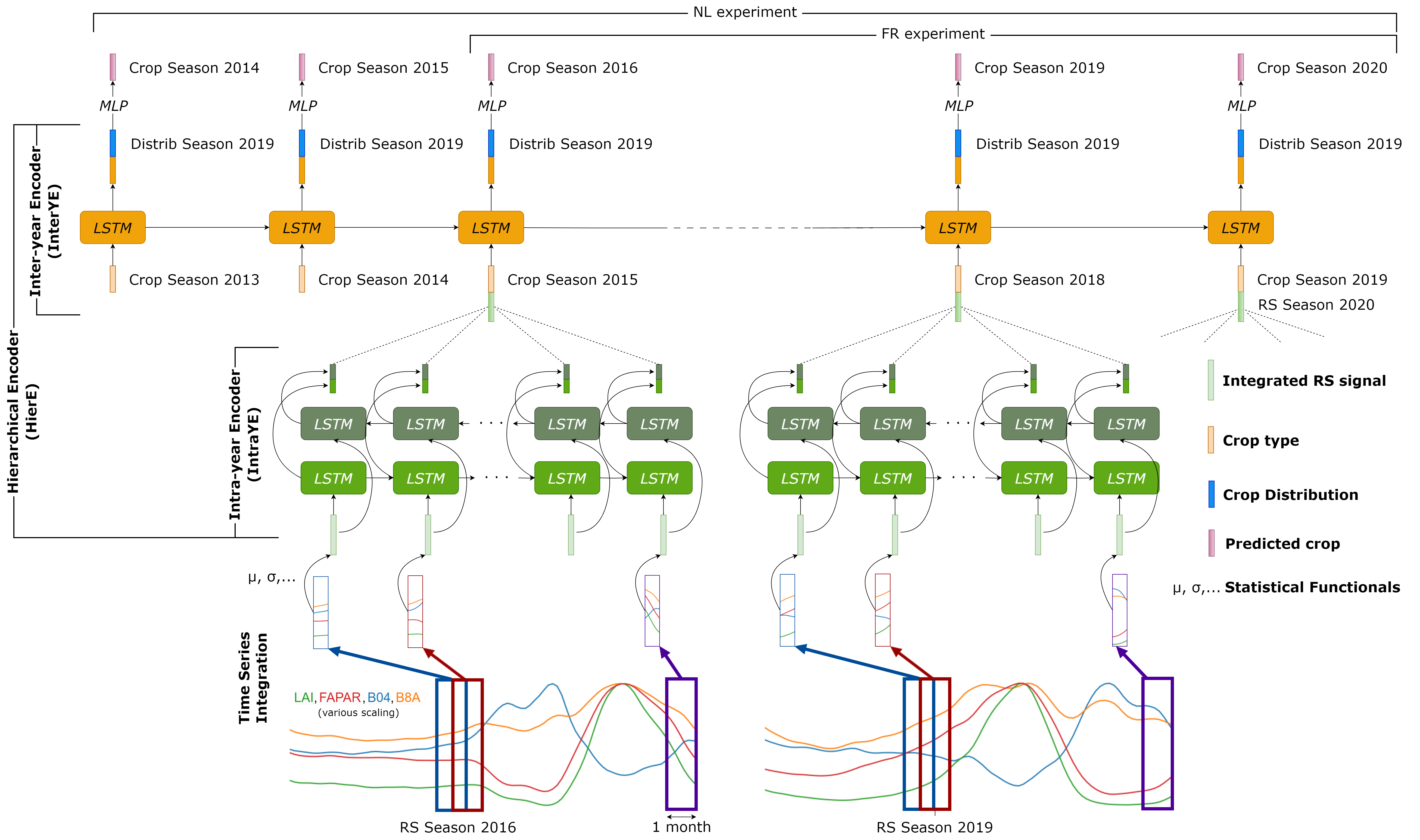} \vspace*{-.5cm}
    \caption{Hierarchical Multimodal Model Conceptual Diagram. The two experiments (NL and FR) are represented with actual seasons used. Crop season 2020 is the final predicted label used for the test set.}
    \label{fig:HierarchicalModel} 
\end{figure*}

As stated in \cite{Russwurm2019}, they only consider the time series of a single season, without incorporating a multi-season modeling approach for the RS data which is a key aspect of our proposed approach.\footnote{This will be shown in next paragraph.}
This unimodal network InterYE$_{RS}$ is the identical component utilized for encoding the \ac{RS} signal at the season-level (one green \ac{IntraYE} in \ref{fig:HierarchicalModel}). This provides a strong \ac{RS} unimodal baseline. The second baseline that we add comes from the work of \cite{Quinton2021}, which is the strongest baseline among the three works incorporating \ac{CR} as per with \ac{RS} data. 
It integrates the \ac{CR} modality by using a one-hot encoder vector of the past crop sequence. In our case, as we model the crops as words, this would mean a well-known simple vector representation for text called Bag-of-Words \citep{Harris1954}, hence we will call it Bag-of-Crops \ac{BoC}. Although this type of representation is known to not work well for short texts like tweets or speech turns \citep{Benamara2016,Barriere2018,Barriere2017b,Barriere2017}, we can expect better results with crop sequences which are considerably less complex than natural language.

\paragraph*{Multi-year non-hierarchical models}
Following the introduction, a set of novel model architectures is suggested hereafter, and their performance is evaluated in comparison to the existing baseline models.
We first aimed to model the sequence of seasons with a recurrent encoder. These models use season-level features (that can be \ac{CR} or \ac{RS}) and are called \ac{InterYE}. This corresponds to the orange top Encoder in \ref{fig:HierarchicalModel}, modeling the sequence of seasons.


We modeled the multi-annual crop rotations in a language model fashion by representing the crops as tokens and learning to predict the next one. This model takes the past sequence of crops $(c_1,...,c_t)$ as inputs and output the new crop $c_{t+1}$, modeling the crop rotation dynamics through the seasons.
This corresponds to the orange \ac{InterYE} in \ref{fig:HierarchicalModel}, if only using crop embeddings. It does not use the blue local crop distribution vector. \textbf{This unimodal model is denoted $\text{InterYE}_{Crop}$}, corresponding to an unimodal Crop Rotations model.

Using solely previous rotations to forecast future crop yields results in inadequate performance due to the limited amount of information provided. Therefore, we decided to enhance the model's robustness by incorporating satellite data, leveraging either the consensus principle or the complementary principle \citep{Xu2013}.
We enhanced the unimodal model $\text{InterYE}_{Crop}$ by adding season-level information from \ac{RS}.
This corresponds to the orange \ac{InterYE} in \ref{fig:HierarchicalModel}, with a green vector being the season-level concatenation of the RS signal (without being processed by the green \ac{IntraYE}). It does not include the blue local crop distribution vector. \textbf{This model is denoted as $\text{InterYE}_{MM}$}, corresponding to a non-hierarchical multimodal model with \ac{RS}. If the model only uses the \ac{RS} modality, then \textbf{it is denoted as $\text{InterYE}_{RS}$}.

\paragraph*{Multi-year hierarchical models}

We chose to model jointly the inter-year and intra-year dynamics with a hierarchical model composed of one network modeling the \ac{RS} dynamics within a season underneath another network modeling the rotation dynamics between the seasons. We processed the \ac{RS} signal beforehand using another \ac{RNN}, before concatenating this unimodal \ac{RS} vector obtained with the crop embedding, in a hierarchical way. 

We incorporated the sequential aspect of the \ac{RS} time-series by processing the \ac{RS} features at the season level with a first sequential encoder before adding their yearly representation into the second neural network modeling the crop types, leading to a hierarchical network with one top network modeling the sequence outputs of another bottom network \citep{Serban2015}.

Compared to InterYE$_{MM}$ and InterYE$_{RS}$, there is another network modeling the RS signal at the bottom. 
This corresponds to the orange color top \ac{InterYE} and the green color \ac{IntraYE} in \ref{fig:HierarchicalModel}, modeling between the seasons as well as within a season. It does not use the blue local crop distribution vector. \textbf{This model is denoted as $\text{HierE}_{MM}$}, corresponding thus to a hierarchical i\ac{IntraYE} and \ac{InterYE} to model both crop rotation and \ac{RS} time-series.

We enhanced the model by adding the \ac{CD} vector after the \ac{IntraYE} because it is a high-level feature regarding the task we are tackling and the deeper you go into the layers the higher-level the representations are w.r.t. the task \citep{Sanh2017}. This corresponds to the full network presented in \ref{fig:HierarchicalModel}, including the blue local crop distribution vector. \textbf{This model is denoted as $\text{HierE}_{final}$}, corresponding thus to a hierarchical \ac{IntraYE} and \ac{InterYE} to model the three modalities.

\paragraph*{Encoders}

We compared several type of models using different architectures and different modalities. Because our work mainly focuses on how to integrate multimodal data, we opted to use \ac{RNN-LSTM}  backbones, proven competitive for this kind of task \citep{Russwurm2020}. Our method is also applicable using other encoders such as transformers \citep{Vaswani2017} or Gated Recurrent Units \citep{Chung2015}. 

For the Inter-year encoder, we first add an embedding layer to transform the crop type $c_t$ at season $t$ into a vector $\textbf{emb}_t = f_e(c_t)$. 
This embedding vector $\textbf{emb}_t$ is used as input of the \ac{LSTM} to produce a hidden state $h_t$ at season $t$  as seen in Equation \ref{eq:rnn}, which will be used to predict the next crop $c_{t+1}$ in Equation \ref{eq:cropmodel}.

\begin{equation}
    \textbf{h}_t = \text{LSTM}_y(\textbf{emb}_t,\textbf{h}_{t-1})
    \label{eq:rnn}
\end{equation}

\begin{equation}
    P(c_{t+1}|c_{t}, ..., c_{1}) = f_c(\textbf{h}_t)
    \label{eq:cropmodel}
\end{equation}

The \ac{RS} features were integrated at the season-level into a feature vector $\textbf{RS}_t$ before the modeling of the crop types by the \ac{LSTM}. 
We feed the season $t$ feature vector $\textbf{RS}_t$ into a neural network layer $f_{rs}$ to reduce its size and then concatenate it with the crop embeddings before the \ac{LSTM} (see Equation \ref{eq:MMRNN}), using $\textbf{emb}_{MM_t}$ instead of $\textbf{emb}_t$ in Equation \ref{eq:rnn}. 
\begin{equation}
    \textbf{emb}_{MM_t} = [\textbf{emb}_t, f_{rs}(\textbf{RS}_t)]
    \label{eq:MMRNN}
\end{equation}

For the \ac{IntraYE}, we chose to use a \ac{biLSTM} with a self-attention mechanism \citep{Bahdanau2016} following the assumption that some parts of the season are more important than others to discriminate the crop type. 
The \ac{biLSTM} is composed of two \ac{LSTM}, one of which reads the sequence forward and the other reads it backward. The final hidden states are a concatenation of the forward and backward hidden states. For a sequence of inputs $[\textbf{RS}_{t_1},...,\textbf{RS}_{t_w}]$ it outputs $w$ hidden states $[\textbf{h}_{RS_{t_1}},..., \textbf{h}_{RS_{t_w}}]$. The self-attention layer\footnote{composed of a feedforward, a relu, another feedforward and a softmax layers} will compute the scalar weights $u_{t_w}$ for each of the $\textbf{h}_{RS_{t_w}}$ (see Equation \ref{eq:att}) in order to aggregate them to obtain the final state $\textbf{h}_{RS_t}$ 
(see Equation \ref{eq:blstm}).

\begin{equation} 
    u_{t_w} = att(\textbf{h}_{RS_{t_w}})
    \label{eq:att}
\end{equation}

\begin{equation} 
    \textbf{h}_{RS_t} = \sum\limits_{w} u_{t_w} \textbf{h}_{RS_{t_w}}
    \label{eq:blstm}
\end{equation}

For the crop distribution, we concatenated the hidden state $\textbf{h}_t$ of the \ac{LSTM} with the crop distribution vector $\textbf{d}$ and mixed them using two fully connected layers $f_{fc1}$ and $f_{fc2}$ (see Equation \ref{eq:distrib}). Hence, we obtain $\textbf{h}_{d_t}$ instead of $\textbf{h}_{t}$ before the final fully connected layer $f_{fc}$ from Equation \ref{eq:cropmodel}.

\begin{equation} 
    \textbf{h}_{d_t} = f_{fc2}(f_{fc1}([\textbf{h}_t, \textbf{d}]))
    \label{eq:distrib}
\end{equation}

\subsection{Automatic Data- and Knowledge-Driven Label Aggregation} 
\label{sec:method_hierchical_aggregation}

\subsubsection{Rationale}
Training and evaluating a model at large scale, on regions that contain different agro-climatic zones is complex due to the heterogeneity of the temporal and spectral representations of the crops and variability of the climate and agricultural practices. 
Indeed, if the labels distribution is highly variable between two datasets, one label that was representative in one domain would become not representative in the other. 
In this work we propose an aggregation of the labels, that would be on the one hand representative of the dataset, and on the other hand thematically pertinent. 
In this way, it should be possible to evaluate the performances of the classification model at different scales: using all the labels from the region, even the ones with very few examples, then using an aggregation of labels that is representative of the region. 
This also offers the advantage to evaluate a model on two different datasets with a relevant evaluation on each dataset. We discuss this method in Subsection \ref{SS:aggregation}.

\subsubsection{A Hierarchical Method to Group Labels}
The labels obtained with the Hierarchical Crop and Agriculture Taxonomy from EuroCrops (see Section \ref{sec:method_study_area}) have heterogeneous distribution and level of interest because of geographically constrained occurrence. We propose a method to merge non-representative crops together to only keep the most relevant in a region of interest by using its label distribution. The method is applied for the evaluation only. 
We take the best of both worlds by fusing expert knowledge and data-driven method. The method is applicable to any dataset, at any geographic scale and fully automatic. 

Benefiting on the hierarchical structure of EuroCrops, we selected crops for which the number of \ac{FOI} is above a given threshold ($th$). The crops with number of \ac{FOI} below $th$ are merged together with the other EuroCrops-sibling crops toward their parent-class. The remaining number of \ac{FOI} falling down into the parent-class is compared with the threshold for an iterative process.



\begin{figure*}[]
	\centering 
	\includegraphics[width=1\textwidth]{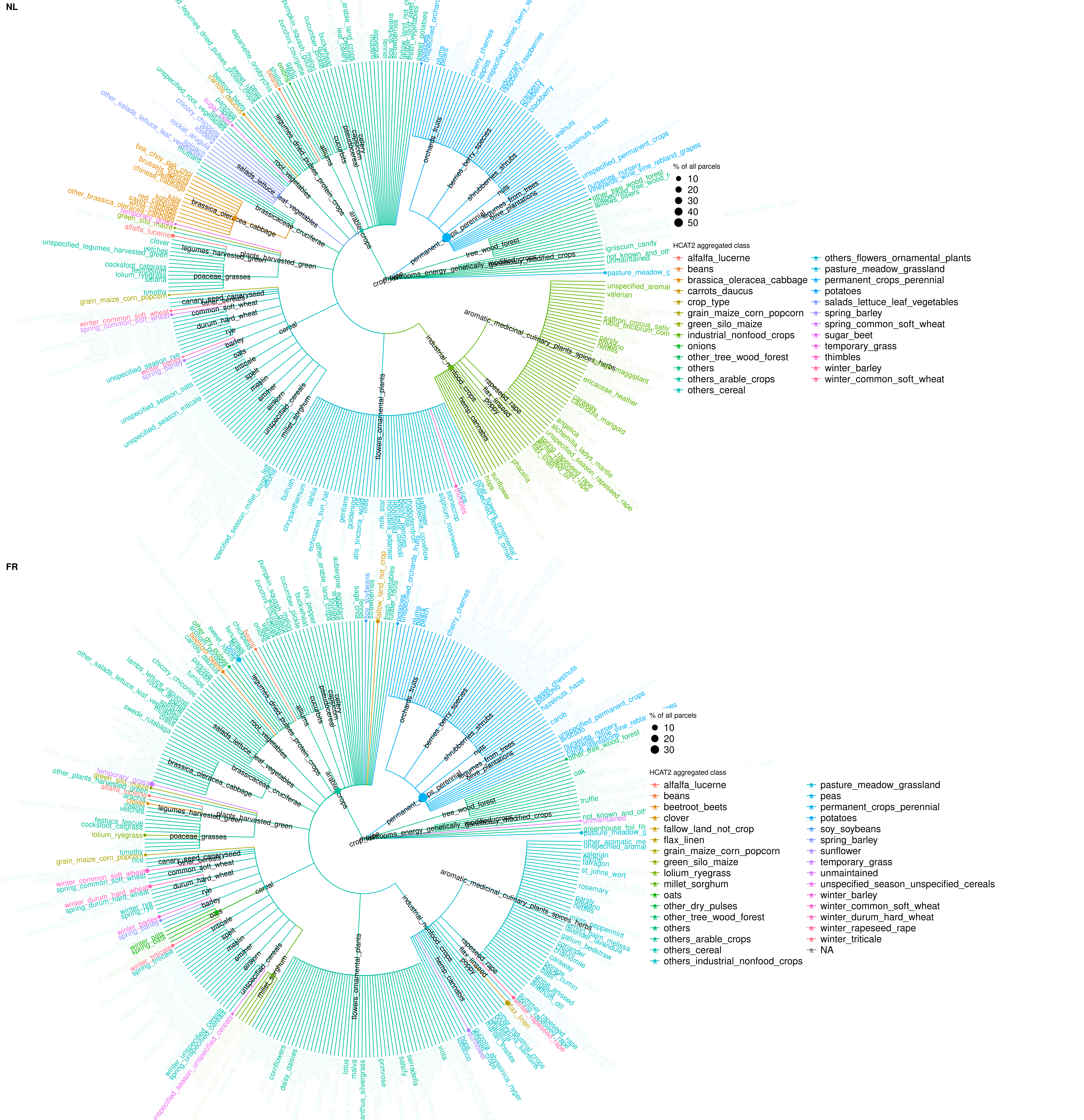} \vspace*{-.5cm}
	\caption{Aggregated classes selected for validation in each country for FR (A) and \ac{NL} (B) along with the distribution. The colour highlight the crop type or crop group that were assessed for both country in red and only for one respective country in cyan.
	}
	\label{fig:Aggregated_Classes_Tree} 
\end{figure*}

For both countries, we set the threshold $th$ at 0.3\% of the dataset size, which roughly corresponds to 2k samples for the \ac{NL} and 20k samples for \ac{FR}. 
We regrouped all the classes falling under the subclass "Permanent Crops" together  
as one for both datasets, as they are the simplest examples to classify when considering rotations. This was done to mitigate its impact by creating multiple labels for permanent crops. 
The automatic aggregation over \ac{FR} and \ac{NL} is shown in \ref{fig:Aggregated_Classes_Tree}.

\subsection{Early-Season data augmentation} 
\label{sec:method_data_augmentation}

Applying the end-of-season models (i.e. trained with the data from the whole season) for early-season (i.e. using incomplete time-series) is not suitable.
We propose a data-augmentation technique in order to help the model to classify a sample even without getting the full time series of the season. The idea behind this method is to force the model to predict the right class even though it does not observe the full time-series, as it would do at end-of-season.

We follow the approach of \cite{Barriere2022cdceo} by randomly cropping the end of the vector feature of the \ac{RS} data $RS = (RS_1, ..., RS_{t_w})$, $t_w$ being the last bi-monthly date of the time-series
varying between 10 and 24.
By setting the minimum number of steps to 10, we ensure that each sample contains sufficient information (at least 5 months) to facilitate the training phase. Knowing the start of the time-series is October, it means we do not crop the end of the time series up to 1st of March. We used the same cropping size $t$ among all the samples of the same mini-batch. For the hierarchical models, we used the same cropping size $t$ among all the seasons.

\subsection{Transfer learning between countries} 
\label{sec:method_transfer_learning}

We ran several experiments in order to take advantage of the normalized taxonomy that we used for both countries, by investigating the potential of transferring knowledge between different domains. For these purposes, we compared the performances of a model trained from scratch (i.e. Vanilla) and a model pre-trained over one country before being fine-tuned over another one. This pre-training allows to transfer knowledge from a source task and domain to a target task and domain.

We tested this approach in (i) cross-domain zero-shot setting, for which the pre-trained model does not see any samples of the new domain, and (ii) cross-domain few-shot setting,\footnote{i.e. when the domain changes as per with significantly new labels} for which the pre-trained model sees limited number samples of the new domain. 
For the cross-domain zero-shot setting, we used a network that was trained over one country on the other country, without fine-tuning it. 
For the cross-domain few-shot setting, we fine-tuned the network only on a subset of the target dataset, taking few examples representative of this dataset. 
We generated the few-shot subsets by randomly sampling $2^N$ (with $N \in \{4,6,8,10\}$) examples of each of the aggregated class (see Section \ref{sec:Feature_Extractions}). 
For this, we sampled using the global crop distribution of 2019, which was used as validation cropping season. We think that this setup is realistic as we only sample from the aggregated classes that are the prominent ones in each of the datasets, and the ones to calculate the metrics to validate the models. 
We added more and more data increasingly so that all the examples from $2^{N_1}$ are comprised in $2^{N_2}$, with $N_1<N_2$. 
A summary of the different experiments can be seen in \ref{tab:TransferLearning}. We did not freeze any layer during the fine-tuning.


\begin{table*}[]
    \footnotesize
    \caption{Settings of the different experiments. $N \in \{4, 6, 8, 10\}$ for the few-shot experiments.} \label{tab:TransferLearning}
    \centering
    \begin{tabular}{@{}lclccc@{}}
    \toprule
    \textbf{Name}       & \textbf{Pre-Training} & \textbf{Training} & \textbf{Testing}  & \textbf{\# data from target}  & \textbf{Models}   \\ \midrule
    few-shot-NL         & $\varnothing$         & NL                & NL                & $2^N$                         & $\text{HierE}_{final}$   \\
    few-shot-FR         & $\varnothing$         & FR                & FR                & $2^N$                         & $\text{HierE}_{final}$   \\ \hline 
    Vanilla-FR          & $\varnothing$         & FR                & FR                & 100\%                         & All models        \\
    Vanilla-NL          & $\varnothing$         & NL                & NL                & 100\%                         & All models        \\ \hline \hline 
    0-shot-NL           & $\varnothing$         & FR                & NL                & 0                             & $\text{HierE}_{final}$   \\
    Transfer-few-shot-NL& FR                    & NL                & NL                & $2^N$                         & $\text{HierE}_{final}$   \\
    transfer-NL         & FR                    & NL                & NL                & 100\%                         & $\text{HierE}_{final}$   \\
    Transfer-FR         & NL                    & FR                & FR                &  100\%                        & $\text{HierE}_{final}$   \\
    \bottomrule
    \end{tabular}
\end{table*}



   

\subsection{Dataset Segmentation and Validation}

The datasets are generally split regarding time or space. \cite{Weilandt2023} proposed to compare the results 
of models trained with or without training data from the same cropping season as the test season, in end-of-season and early-season settings. They found out that the difference in performance was minimal. 

In this work, we are interested to apply early-season setting and therefore segmenting the train and test sets among seasons. This provides two key advantages: we are in a real-life setting without in-situ data from the test season, and prohibit training with data from the end of the season in early-season setting. 
We trained our networks as for a sequence classification task, always with several seasons of data. The labels up to 2018 were used as training set, while the labels from 2019 were used as validation set and the labels from 2020 as test set (see \ref{fig:HierarchicalModel}). All results presented hereafter refer to the analysis of 2020 crop types, which are based on models trained with the period 2013-2019 for \ac{NL} and 2015-2019 for \ac{FR}, thus independent from the 2020 crop types observations. We zero-padded when no RS data was available (before 2016).

In order to confirm \cite{Weilandt2023} findings, we added experiments with a temporal and spatial split of the data in Appendix \ref{AppendixB}, showing that the results of a model trained on other seasons and other parcels still reach high results. 

We validated the models using metrics calculated for four level of aggregation: (i) with all the labels, (ii) with the aggregation of the labels using the Automatic Hierarchical Label Aggregation, (iii) with a set of crop of interest from the aggregation that were recognized important by a Food Security expert and with or (iv) without classes \textit{others} and \textit{grassland} (majority class in \ac{NL} dataset). We used macro-average of the Precision, Recall and F1-score because the dataset is imbalanced. We also used \ac{m-F1} for the last setting, because accuracy is not possible when focusing on a subset of the classes. In addition, we computed the Accuracy as a general metric.

\subsection{Implementation} 


We trained all the networks via mini-batch stochastic gradient descent using the Adam optimizer \citep{Kingma2014} with a learning rate of $10^{-3}$ and a cross-entropy loss function. The number of neurons for the crop embedding layer,\footnote{from 16 to 128, best 64} both the \ac{RNN} internal layers,\footnote{from 64 to 512, best 256} and the fully connected \ac{RS} layer $f_{rs}$\footnote{from 32 to 256, best 128} as well as the number of stacked \ac{LSTM}\footnote{from 1 to 4, best 3} were chosen using hyperparameters grid search of power of 2 on a subset of the \ac{NL} dataset. The sizes of the layers $f_{c1}$ and $f_{c2}$ are the same as the one from the second \ac{RNN} state $\textbf{h}_t$.

\subsection{Processing facility, data and code}
\label{sec:method_processing}

The \ac{EO} extraction and processing, the classification and the benchmarking were performed on the JRC Big \ac{BDAP} using an HTCondor environment \citep{soille2018versatile}. The platform\footnote{\url{https://jeodpp.jrc.ec.europa.eu/bdap/}} has been built upon the near-data processing concept, which prescribes placing the computing facility close to the storage units to avoid the bottleneck of delaying or degrading interconnection. 
Experiments with the neural networks were run using PyTorch 1.4.0 \citep{Paszke2019} on a GPU Nvidia RTX-8000 using CUDA 12.0. The training phase allows to process between roughly 5k and 30k examples per seconds, with each example containing 4 seasons of data.

The data extracted and used for this study are openly available on the public FTP\footnote{ The data can be downloaded on \url{https://jeodpp.jrc.ec.europa.eu/ftp/jrc-opendata/DRLL/CropDeepTrans/}.}. The code for the data processing, the labels aggregation, and deep learning experiments will be freely available after publication.\footnote{Add URL after publication}


\section{Experiments and Results} \label{sec:experiments}

\subsection{Feature Extractions}
\label{sec:Feature_Extractions}
\subsubsection{Crop Rotation}

\begin{figure*}[]
    \centering 
    \includegraphics[width=1.\textwidth]{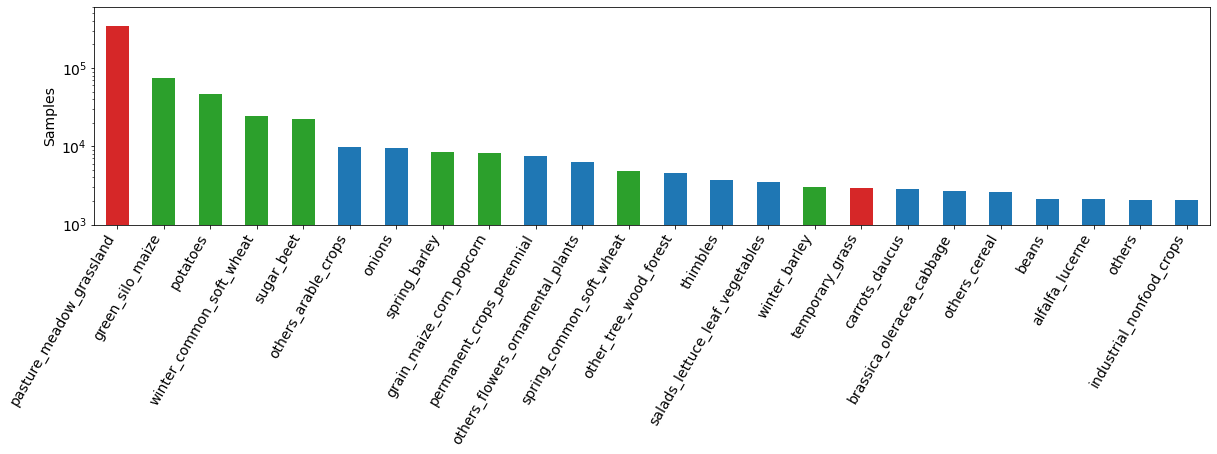} \vspace*{-.5cm}
    \caption{Distributions of the crop types in the \ac{NL} dataset for the test season (2020), after aggregation. Green bars are the selected crops for the 8-class evaluation; Red bars refer to the grasslands, and blue bars to the remaining crops.}
    \label{fig:Distrib_NL} \vspace*{-.5cm}
\end{figure*}

\begin{figure*}[]
    \centering 
    \includegraphics[width=1.\textwidth]{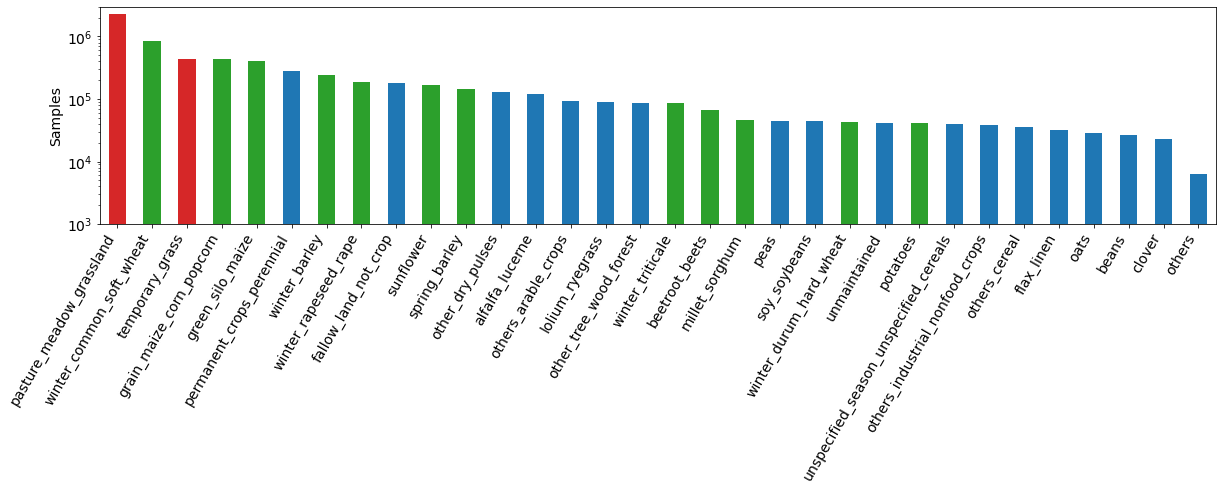} \vspace*{-.5cm}
    \caption{Distributions of the crop types in the France dataset for the test season (2020), after aggregation. Green bars are the selected crops for the 12-class evaluation; Red bars refer to the grasslands, and blue bars to the remaining crops.}
    \label{fig:Distrib_FR} \vspace*{-.5cm}
\end{figure*}

The crop label categories for the 2020 test set correspond to a long-tailed class distributions, as shown for the 32-class and 24-class aggregations for the French and the Dutch data sets in \ref{fig:Distrib_NL} and \ref{fig:Distrib_FR}, respectively. The models are finally validated on a set of crops of interest from the 32-class and 24-class aggregation. Those 8- and 12-class, respectively for the \ac{NL} and \ac{FR}, were identified by experts as essential for crop production monitoring and food security. They are shown in green in \ref{fig:Distrib_NL} and \ref{fig:Distrib_FR}.

\subsubsection{RS-based Features} 

First, the \ac{RS} data retrieved as described in \ref{sec:method_eo} were obtained. The final dataset which consists in the full time series of more than 7M \ac{FOI}, for a total of more than 35M FOI-season (NL : 5 seasons x 596k FOI; FR: 5 seasons x 6,49M) are available for download, as well as the extracted features used in the experiments\footnote{\url{https://jeodpp.jrc.ec.europa.eu/ftp/jrc-opendata/DRLL/CropDeepTrans/}}.

\begin{table*}[!ht]
	\centering
	\resizebox{1\textwidth}{!}{
		\begin{tabular}{l|c|llll|llll|llll|llll}
			\textbf{Labels} & \multirow{2}{*}{\textbf{\# Modalities}}  &  \multicolumn{4}{c}{\textbf{141-class}} & \multicolumn{4}{c}{\textbf{24-class}} & \multicolumn{4}{c}{\textbf{10-class}} & \multicolumn{4}{c}{\textbf{8-class}} \\ 
			\textbf{Model} & & P & R & F1 & Acc &  P & R & F1 & Acc &  P & R & F1 & Acc &  P & R & F1 & m-F1 \\  \hline \hline 
			$\text{InterYE}_{Crop}$    &1   (C)     & 36.0 & 25.5 & 27.4 & 76.2 & 53.3 & 37.2 & 39.1 & 76.5 & 51.8 & 43.0 & 43.5 & 77.7 & 43.3 & 35.5 & 34.9 & 53.6 \\\hline 
			$\text{IntraYE}_{RS}$ \citeyear{Russwurm2019}   & 1 (RS)        & 27.4 & 20.9 & 20.4 & 89.8 & 64.0 & 60.9 & 60.4 & 90.3 & 78.8 & 75.9 & 74.5 & 92.9 & 76.1 & 72.6 & 70.8 & 87.8 \\
			$\text{InterYE}_{RS}$    &1     (RS)      &  22.8 & 17.7 & 17.1 & 89.1 & 59.2 & 58.5 & 57.3 & 89.6 & 71.2 & 73.4 & 72.0 & 92.1 & 67.0 & 69.6 & 68.0 & 85.6 \\ 
			$\text{HierE}_{RS}$   &1    (RS)        &  20.7 & 17.5 & 16.7 & 90.2 & 64.3 & 61.0 & 61.2 & 90.9 & 80.5 & 74.4 & 74.3 & 93.5 & 78.0 & 70.4 & 70.3 & 88.3 \\ \hline
			$\text{IntraYE}_{MM}$ \citeyear{Quinton2021} &2  (RS+BoC)  & 55.6 & 39.7 & 43.2 & 92.8 & 76.6 & 69.8 & 72.1 & 93.1 & 83.0 & 80.5 & 80.9 & 94.7 & 80.2 & 77.9 & 78.0 & 90.0\\
			$\text{InterYE}_{MM}$   &2  (RS+C)  & 41.1 & 33.0 & 33.6 & 92.2 & 70.8 & 70.5 & 69.9 & 92.6 & 82.2 & 79.7 & 80.4 & 94.5 & 80.2 & 76.3 & 77.5& 89.5 \\
			$\text{HierE}_{MM}$ & 2 (RS+C)& 47.3 & 38.7 & 39.7 & 93.3 & 74.7 & 75.5 & 74.7 & 93.7 & 85.2 & 81.9 & 83.1 & 95.2 & 83.6 & 78.8 & 80.6 & 91.1 \\
			$\text{HierE}_{final}$ & 3 (All)&  47.1 & 39.3 & 40.2 & \textbf{93.6} & 76.6 & 75.8 & 75.7 & \textbf{94.0} & 86.7 & 81.9 & 83.6 & \textbf{95.5} & 85.3 & 78.7 & 81.1 & \textbf{91.6} \\
		\end{tabular}
	}
	\caption{Results over Netherlands of the end-of-season classification models with different modalities: Remote Sensing (RS), Crop Rotations as embeddings (C) or \ac{BoC}, and Spatial Crop Distribution. 
		The metrics shown are macro Precision (P), Recall (R) and F1 score, as well as accuracy and micro-F1 score (m-F1). 
	}
	\label{tab:Results_NL}
	\vspace*{-.4cm}
\end{table*}

\begin{table*}[!ht]
	\centering
	\resizebox{1\textwidth}{!}{
		\begin{tabular}{l|c|llll|llll|llll|llll}
			\textbf{Labels} & \multirow{2}{*}{\textbf{\# Modalities}}  &  \multicolumn{4}{c}{\textbf{151-class}} & \multicolumn{4}{c}{\textbf{32-class}} & \multicolumn{4}{c}{\textbf{14-class}} & \multicolumn{4}{c}{\textbf{12-class}} \\ 
			\textbf{Model} & & P & R & F1 & Acc &  P & R & F1 & Acc &  P & R & F1 & Acc &  P & R & F1 & m-F1 \\  \hline \hline 
			$\text{InterYE}_{Crop}$    &1   (C)     & 35.6 & 31.0 & 31.7 & 66.0 & 43.7 & 38.8 & 38.7 & 66.2 & 38.9 & 34.3 & 31.7 & 69.1 & 30.9 & 26.4 & 23.0 & 42.7 \\\hline
			$\text{IntraYE}_{RS}$ \citeyear{Russwurm2019}   & 1 (RS)        & 22.9 & 15.7 & 15.2 & 64.0 & 51.1 & 46.0 & 44.6 & 64.5 & 69.8 & 62.2 & 64.7 & 75.7 & 69.3 & 59.7 & 63.1 & 74.6 \\
			$\text{InterYE}_{RS}$    &1     (RS)      &  21.3 & 13.2 & 12.6 & 54.9 & 46.5 & 41.5 & 39.2 & 55.5 & 63.9 & 59.6 & 60.2 & 72.2 & 62.7 & 57.4 & 58.5 & 71.2\\ 
			$\text{HierE}_{RS}$   &1    (RS)        &  25.3 & 19.0 & 18.8 & 66.3 & 55.5 & 50.7 & 50.1 & 66.9 & 72.5 & 65.5 & 67.7 & 76.9 & 71.9 & 63.2 & 66.1 & 76.5\\ \hline
			$\text{IntraYE}_{MM}$ \citeyear{Quinton2021} &2  (RS+BoC)  & 52.7 & 32.4 & 35.9 & 82.7 & 70.1 & 59.3 & 61.8 & 82.8 & 78.1 & 68.7 & 71.0 & 86.6 & 76.2 & 65.6 & 68.0 & 80.3\\
			$\text{InterYE}_{MM}$   &2  (RS+C)  & 45.9 & 35.2 & 36.4 & 82.4 & 67.7 & 60.5 & 62.4 & 82.7 & 72.7 & 67.4 & 69.2 & 86.1 & 70.0 & 63.6 & 65.8 & 77.5\\
			$\text{HierE}_{MM}$ & 2 (RS+C)& 50.2 & 41.9 & 43.2 & 84.8 & 70.7 & 67.6 & 68.2 & 85.0 & 77.0 & 73.4 & 74.9 & 88.4 & 75.0 & 70.2 & 72.3 & 81.8\\
			$\text{HierE}_{final}$ & 3 (All)&  45.1 & 37.3 & 38.1 & \textbf{85.4} & 72.1 & 68.8 & 69.2 & \textbf{85.7} & 79.8 & 76.1 & 77.6 & \textbf{89.1} & 78.1 & 73.5 & 75.4 & \textbf{83.6}\\ 
		\end{tabular}
	}
	\caption{Results over France of the end-of-season classification models with different modalities: Remote Sensing (RS), Crop Rotations as embeddings (C) or Bag-of-Crops (BoC), and Spatial Crop Distribution. 
		The metrics shown are macro precision, recall and F1 score, as well as accuracy and micro-F1 score (m-F1).
	}
	\label{tab:Results_FR}
	\vspace*{-.4cm}
\end{table*}

\begin{figure*}[!ht]
	\hspace*{-1.9cm}
	\includegraphics[width=1.2\textwidth]{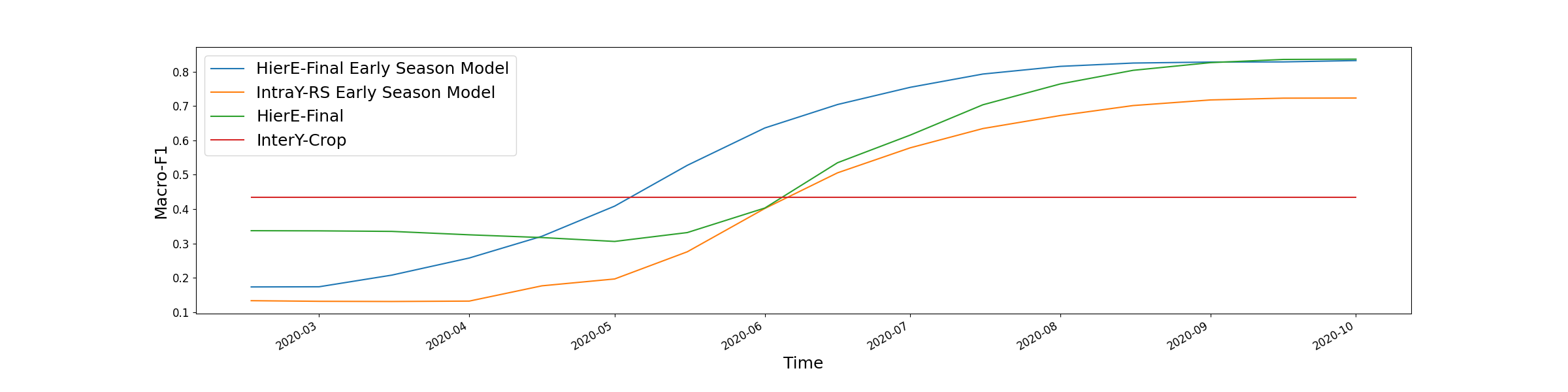} 
	\caption{Comparison of early classification using different modalities, with/out data augmentation (macro-F1 with 10 classes) on Netherlands.} 
	\label{fig:MF1_ModalitiesTimeSerie_NL} 
\end{figure*} 

\subsection{In-country end-of-season results} 

The results of the various models on France and the \ac{NL} are shown in \ref{tab:Results_NL} and \ref{tab:Results_FR} with respect to four distinct classification schemes, ranging from a fine-grained 141-class scheme for \ac{NL} (resp. 151 for \ac{FR}) to a coarse-grained 10-class scheme (resp. 14 for \ac{FR}). The 8-class scheme (resp. 12 for \ac{FR}) is the same than the 10-class one, in order to only focus the performances on the crops of interest. 
\ref{tab:Results_NL} and \ref{tab:Results_FR} aim at showing the interest of our multi-modal method compared to what is generally used in the field, by using only the remote sensing of the season, in an independent way. 


\subsection{In-country early-season results}


We compared the performance of our best model trained without the data-augmentation technique, i.e. trained solely on end-of-season classification examples, with an in-season model of the same architecture trained using the proposed data-augmentation technique (see Section \ref{sec:method_data_augmentation}).
Comparisons were also made with a unimodal model using only the \ac{CR}, and an unimodal IntraYE$_{RS}$ enhanced with the data-augmentation technique.  

\ref{fig:MF1_ModalitiesTimeSerie_NL} shows the performances of the model in terms of micro-F1 on the set of 10 crops. For sake of clarity, we focused on the results of the model trained using this data-augmentation over the \ac{NL}.

\subsection{Cross-country transfer learning results}

\begin{table*}[]
	\centering
	\hspace*{-.5cm}
	\resizebox{1\textwidth}{!}{
		\begin{tabular}{c|c|llll|llll|llll|llll}
			\textbf{Labels} & \multirow{2}{*}{\textbf{N}}  &  \multicolumn{4}{c}{\textbf{141-class}} & \multicolumn{4}{c}{\textbf{24-class}} & \multicolumn{4}{c}{\textbf{10-class}} & \multicolumn{4}{c}{\textbf{8-class}} \\ 
			\textbf{Pre-train.} & & P & R & F1 & Acc &  P & R & F1 & Acc &  P & R & F1 & Acc &  P & R & F1 & m-F1 \\  \hline \hline 
			\multirow{5}{*}{\xmark} &  0    & $\varnothing$ & $\varnothing$ & $\varnothing$ & $\varnothing$ & $\varnothing$ & $\varnothing$ & $\varnothing$ & $\varnothing$ & $\varnothing$ & $\varnothing$ & $\varnothing$ & $\varnothing$ & $\varnothing$ & $\varnothing$ & $\varnothing$ & $\varnothing$ \\
			&  16    & 5.8 & 5.1 & 4.8 & 70.8 & 23.7 & 21.4 & 20.4 & 71.1 & 38.5 & 37.4 & 36.3 & 73.6 & 38.5 & 37.4 & 36.3 & 45.3 \\
			& 64     & 2.7 & 2.5 & 2.2 & 69.2 & 17.1 & 13.1 & 12.5 & 69.4 & 27.3 & 25.7 & 23.3 & 69.6 & 27.3 & 25.7 & 23.3 & 34.7 \\
			& 256   &  4.2 & 4.8 & 2.9 & 66.5 & 18.2 & 16.9 & 14.1 & 66.8 & 25.0 & 23.2 & 20.5 & 68.1 & 25.0 & 23.2 & 20.5 & 20.4 \\ 
			& 1024      &  19.6 & 13.3 & 12.4 & 80.8 & 53.6 & 39.8 & 37.2 & 80.3 & 69.7 & 60.4 & 61.5 & 84.0 & 69.7 & 60.4 & 61.5 & 76.3\\ 
			\hline \hline
			\multirow{5}{*}{\cmark} &  0    & 5.7 & 4.8 & 4.2 & 47.3 & 14.7 & 15.1 & 11.1 & 46.6 & 20.6 & 19.7 & 16.6 & 46.9 & 12.3 & 7.4 & 8.4 & 24.5 \\
			& 16    & 12.2 & 7.8 & 7.6 & 70.3 & 30.5 & 23.8 & 24.5 & 70.4 & 37.9 & 33.9 & 34.0 & 72.3 & 37.9 & 33.9 & 34.0 & 45.2 \\
			& 64     & 16.7 & 13.6 & 13.5 & 74.7 & 41.9 & 38.7 & 38.1 & 75.0 & 51.6 & 45.4 & 46.6 & 76.4 & 51.6 & 45.4 & 46.6 & 54.4\\
			& 256   &  25.8 & 21.4 & 20.8 & 82.5 & 55.6 & 51.1 & 50.6 & 82.7 & 67.3 & 58.0 & 60.1 & 84.6 & 67.3 & 58.0 & 60.1 & 69.2 \\ 
			& 1024      &  32.7 & 27.3 & 26.0 & 84.9 & 61.3 & 57.3 & 54.3 & 84.9 & 73.8 & 72.0 & 71.6 & 87.0 & 73.8 & 72.0 & 71.6 & 80.9\\ 
			\hline \hline
			\xmark & All &  47.1 & 39.2 & 40.2 & 93.7 & 76.6 & 75.8 & 75.8 & 94.0 & 86.7 & 81.9 & 83.6 & 95.5 & 85.3 & 78.7 & 81.1 & 91.6 \\
			\cmark & All &  42.5 & 35.3 & 36.0 & 92.8 & 67.3 & 53.4 & 55.9 & 94.2 & 89.9 & 82.2 & 85.3 & 95.7 & 88.8 & 77.6 & 82.3 & 91.8 
		\end{tabular}
	}
	\caption{Results over Netherlands of the few-shot final classification models, with or without pre-training over France. N represent the number of examples shown per aggregated class on the target dataset. 
		The metrics shown are macro precision, recall and F1 score, as well as accuracy and micro-F1 score (m-F1).
		N stands for the Few-Shot size.} \vspace*{-.5cm}
	\label{tab:FewShots_NL}
\end{table*}

\ref{tab:FewShots_NL} shows the performance of our best architecture model on the \ac{NL}. The neural network was trained either from scratch or using a pre-trained model. We compared its performances on a few-shot setting against using the full \ac{NL} dataset.  
The table shows that pre-training the model led to improved performance in terms of almost in terms of all metrics for all tasks. Furthermore, the performance increases with the number of training examples, and the highest performance was achieved for 1024 training examples for each aggregated class. Notice that the non-pretrained model (i.e., red-cross) using $N=All$ corresponds to $\text{HierE}_{final}$ from \ref{tab:Results_NL}.

\section{Discussion} \label{S:Discussion}

The results are discussed in the five first subsections of the section, then followed by a presentation of the limitations (Section \ref{S:limitations}) and recommendations for future research (section \ref{sec:recomend}).

\subsection{The label aggregation method} \label{SS:aggregation}

In line with second objective, we introduced a distinct benchmarking approach that utilizes HCATv2, the hierarchical crop type classification system of EuroCrops. 
This is a method to output pertinent metrics on a dataset that has many classes with a long tail distribution, by grouping together the minoritarian classes that are similar (i.e. close in the HCATv2 graph). 
In \ref{tab:Results_NL} (NL) and \ref{tab:Results_FR} (FR), the aggregation level is displayed in the 24-class and 32-class columns, respectively. This means that the performance results for both countries incorporate the internal distributions. Overall, the accuracy scores for both countries did not exhibit significant variation when accounting for all EuroCrops classes (93.6 and 85.4, for \ac{NL} and FR, respectively) or after aggregation (94.0 and 85.7), suggesting that the models performed well on the very dominant classes. However, the macro F1-scores experienced an improvement due to the merging of classes with a limited number of samples: it reveals the interest of the aggregation method for classifier performances evaluation on a dataset with a high number of crop classes.

\subsection{Hierarchical multimodal models: a way to gain in performances}

The third objective of the study (section \ref{sec:objectives}) is to evaluate the performances of models relying on diverse model configurations involving different modalities. 

For the \ac{NL}, the results of \ref{tab:Results_NL} indicate that the best performance was achieved by the final model ($\text{HierE}_{final}$) combining the three modalities. It achieved a macro-F1 score of 40.2\% on the 141-class scheme and a micro-F1 score of 91.6\% on the 8-class scheme. 
With the exception of $\text{HierE}_{RS}$, which failed to learn during the optimization process even on the training set (achieving an F1 score of 0 for both the 8- and 12-class settings), all models based on \ac{LSTM} to process RS time series achieve higher performances compared to the approach of concatenating statistical functional vectors of each sliding window. Notably, the unimodal $\text{IntraYE}_{RS}$ outperformed $\text{InterYE}_{RS}$, while the multimodal $\text{IntraYE}_{MM}$ and $\text{HierE}_{MM}$ outperform $\text{InterYE}_{MM}$. More details are found in the next subsection (\ref{SS:percrops}).

Several similarities and notable differences can be observed when comparing the models between the \ac{NL} (\ref{tab:Results_NL}) and \ac{FR} (\ref{tab:Results_FR}). 
First, the general performance metrics were lower for \ac{FR} than for the \ac{NL}. This difference may be attributed to \ac{FR} having more classes than the \ac{NL}.
 Second, the performances of the RS model's performance using only one season of context ($\text{InterYE}_{RS}$) were significantly lower than for the same one on NL, relatively to the other models. The higher variance of the RS data in \ac{FR}, resulting from its larger size and greater diversity compared to the \ac{NL}, may contribute to the lower performances observed in this scenario.

\subsection{Performances per crops} \label{SS:percrops}

\begin{figure*}[]
    \centering \hspace*{-1cm}
    \includegraphics[width=1\textwidth]{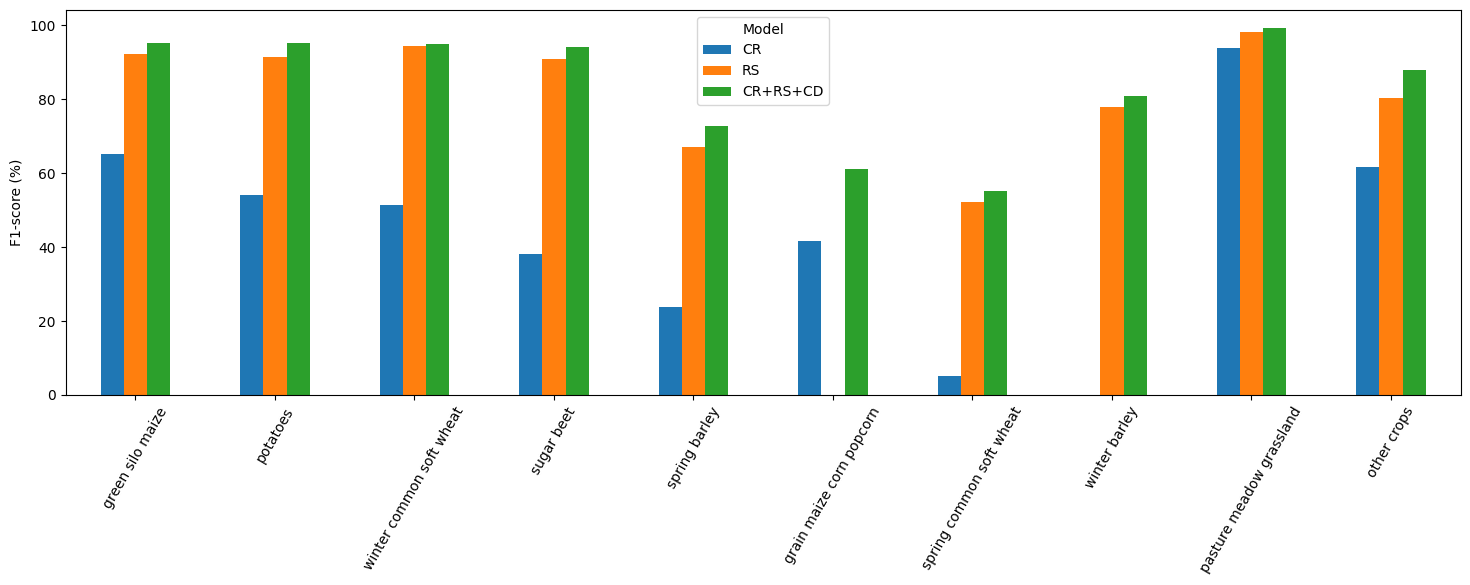} 
    \caption{Comparison of the F1-scores by crops of the best hierarchical multimodal model and the model using different modalities (Crop rotation, Remote sensing only and all) on the Netherlands. We used the $\text{InterYE}_{Crop}$ (in blue), $\text{IntraYE}_{RS}$ (in orange) and $\text{HierE}_{final}$  (in green) models.}
    \label{fig:Modalities_NL}
\end{figure*}

\begin{figure*}[]
    \centering \hspace*{-1cm}
    \includegraphics[width=1\textwidth]{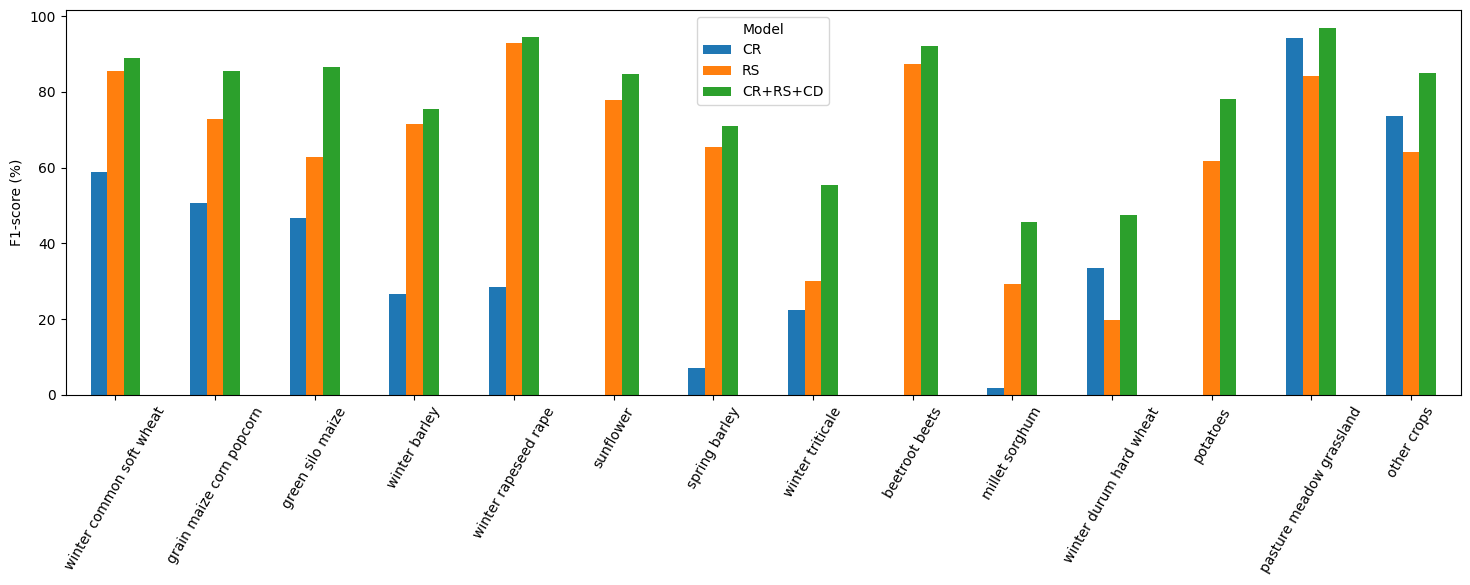} 
    \caption{Comparison of the F1-scores by crops of the best hierarchical multimodal model and the model using different modalities (Crop rotation, Remote sensing only and all) 
    on France 
    We used the $\text{InterYE}_{Crop}$, $\text{IntraYE}_{RS}$ and $\text{HierE}_{final}$ models.}
    \label{fig:F1_Modalities_FR}
\end{figure*}

The utilization of various modalities and their combinations for each of the primary crops in both countries is depicted in \ref{fig:Modalities_NL} and \ref{fig:F1_Modalities_FR}. It demonstrated an upward trend in the level of improvement with an increase in the number of modalities employed. The benefits of crop rotation were more significant for certain crops such as pasture, while others such as beetroot are harder to predict without RS signal. 
This suggested that the crop rotation for \ac{FR}, limited to only starting from 2015, might not offer enough information to accurately predict crops in complex or irregular crop rotation sequences.

\begin{figure*}[]
    \hspace*{-.5cm}
    \centering 
    \includegraphics[width=1.08\textwidth]{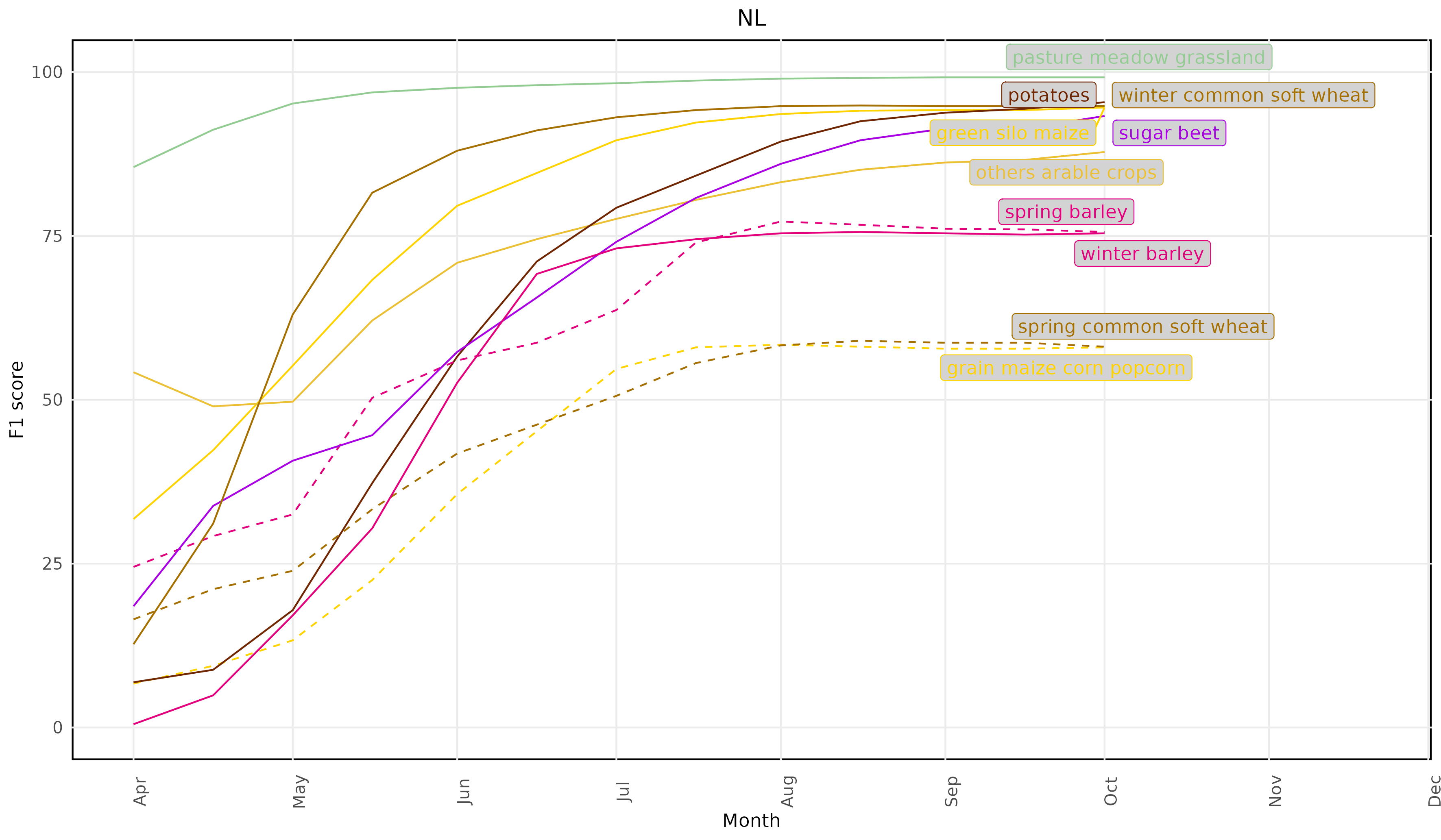}
    \caption{F1-score for each crop group along the season for \ac{NL}. When crop have winter and spring varieties, the spring varieties are represented as dashed lines.} \vspace*{-.5cm}
    \label{fig:F1_Crops_NL} 
\end{figure*}

\ref{fig:F1_Crops_NL} provides more detailed information on the performance of the best model in the \ac{NL}. It displays the F1 scores for eight crops of interest in the \ac{NL}, based on the time of the season used for prediction. Furthermore, for the same crops, we analysed the time series data for the 2020 cropping season (i.e., the test season) averaged at the country level for each of the four remotely sensed variables on \ref{fig:Mean_Crop_EO_timeSeries}, accompanied by the standard error shown on \ref{fig:STD_Crop_EO_timeSeries}. These visualizations highlight both the variability between crops and the potential confusion that can arise between different crop types. As the season progresses, there is a noticeable trend of increasing F1-scores, highlighting the improved accuracy of crop discrimination. However, there are evident disparities among crops. Notably, grassland demonstrates a high F1-score at the season's onset and maintains this level from the end of spring onwards. This is attributed to the fact that the unique temporal signature of grassland phenology doesn't provide significant discriminatory power as the season advances. Contrarily, crops with distinct morphological and phenological characteristics, such as potatoes, green silo maize, and sugar beet, achieve higher F1-scores earlier in the season. The F1-score of barley reaches a peak of 0.75 by the end of July. An intriguing observation is that during the early spring, spring barley exhibits a higher F1-score. However, by June, winter barley surpasses it. This shift can be elucidated by the differential crop calendars that spring and winter barley adhere to.

\begin{figure*}[]
    \centering \hspace*{-1cm}
    \includegraphics[width=1\textwidth]{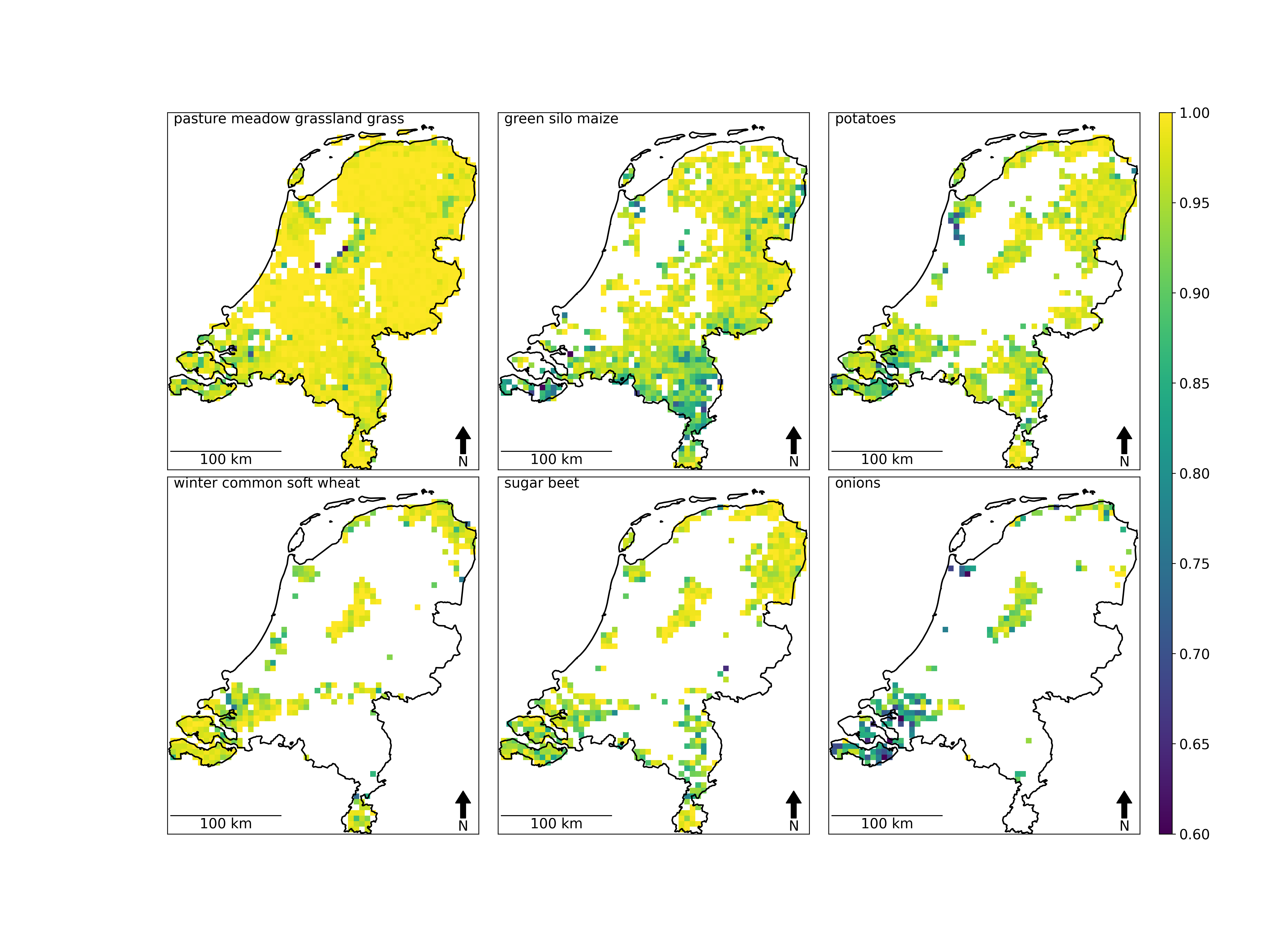}
    \caption{Map of F1 for the six most important crops over The Netherlands. The F1-score is computed for each crop and for each 5km grid cell. Grid cells with less than 50ha (i.e., 2\% of the land) of the given crops are not plotted. Map projection is EPSG:3035. 
    }
    \label{fig:F1_Map_NL}
\end{figure*}

\begin{figure*}[]
    \centering \hspace*{-1cm}
    \includegraphics[width=1\textwidth]{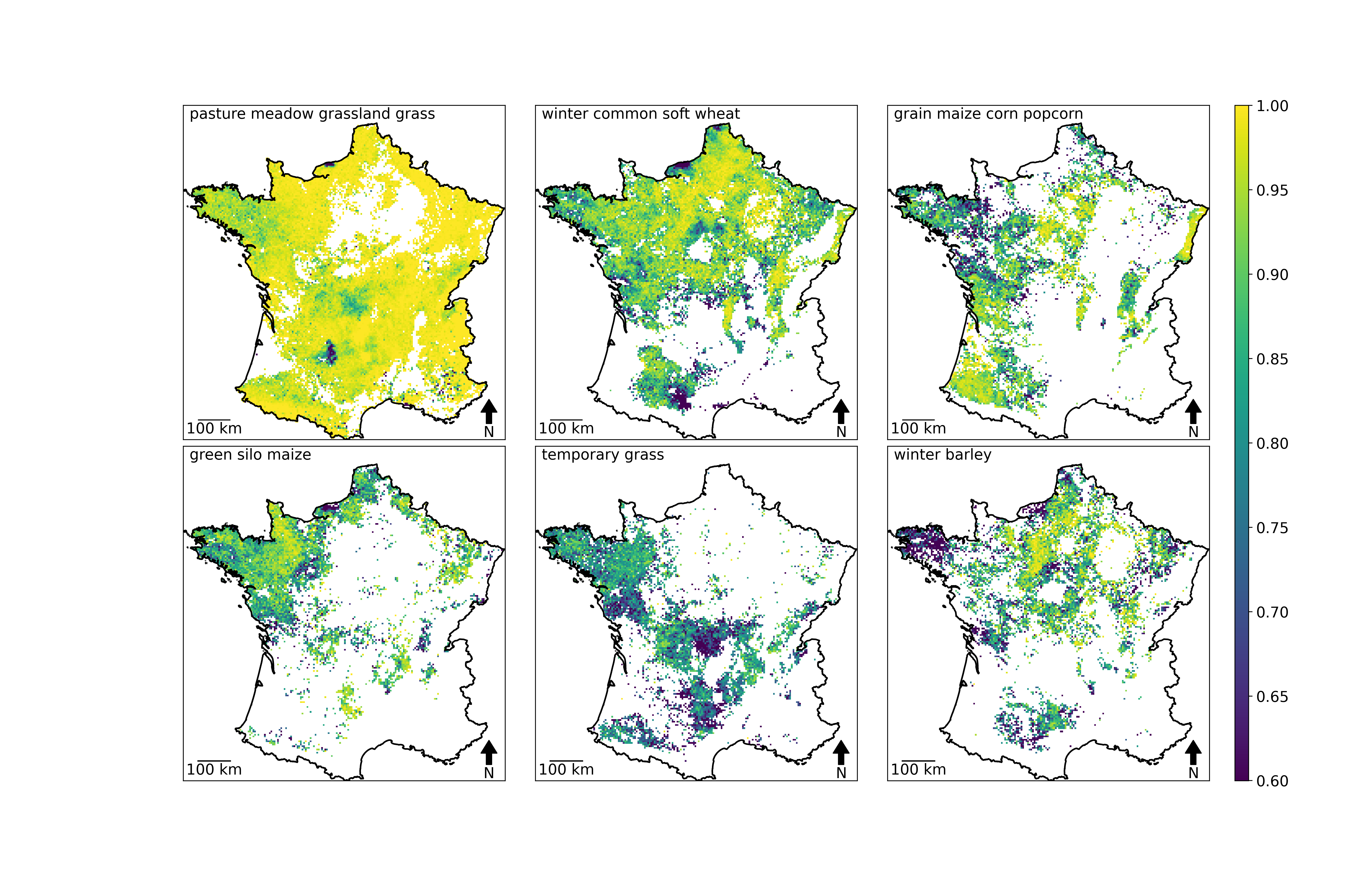}
    \caption{Map of F1 for the six most important crops over Metropolitan France. See legend of \ref{fig:F1_Map_NL}.}
    \label{fig:F1_Map_FR}
\end{figure*}

The end-of-season performances of the models in the \ac{NL} and FR, denoted as $\text{HierE}_{final}$, are presented using 5km grid cell maps in \ref{fig:F1_Map_NL} and \ref{fig:F1_Map_FR} respectively. The maps reveal notable regional effects, particularly in \ac{FR}. For instance, in Brittany (located in the north-west of \ac{FR}), lower accuracy was observed for most crops, especially \textit{winter barley}. It worth noting that despite labeling the crop types uniformly by country, variations in crop varieties, climates, and agricultural practices among farmers have an impact on phenology, consequently affecting the RS signal. Consequently, using a single dataset for training the model resulted in heterogeneous performances over a large country like \ac{FR}. The development of regional models is thus highly recommended.
Furthermore, the model lacks information regarding the specific locations, except through the proxy of crop distribution. Incorporating geographic coordinates and/or weather variables as model inputs could contribute to accounting for spatial variations over large areas, such as \ac{FR}.

\begin{figure*}[]
    \centering \hspace*{-1cm}
    \includegraphics[width=1.1\textwidth]{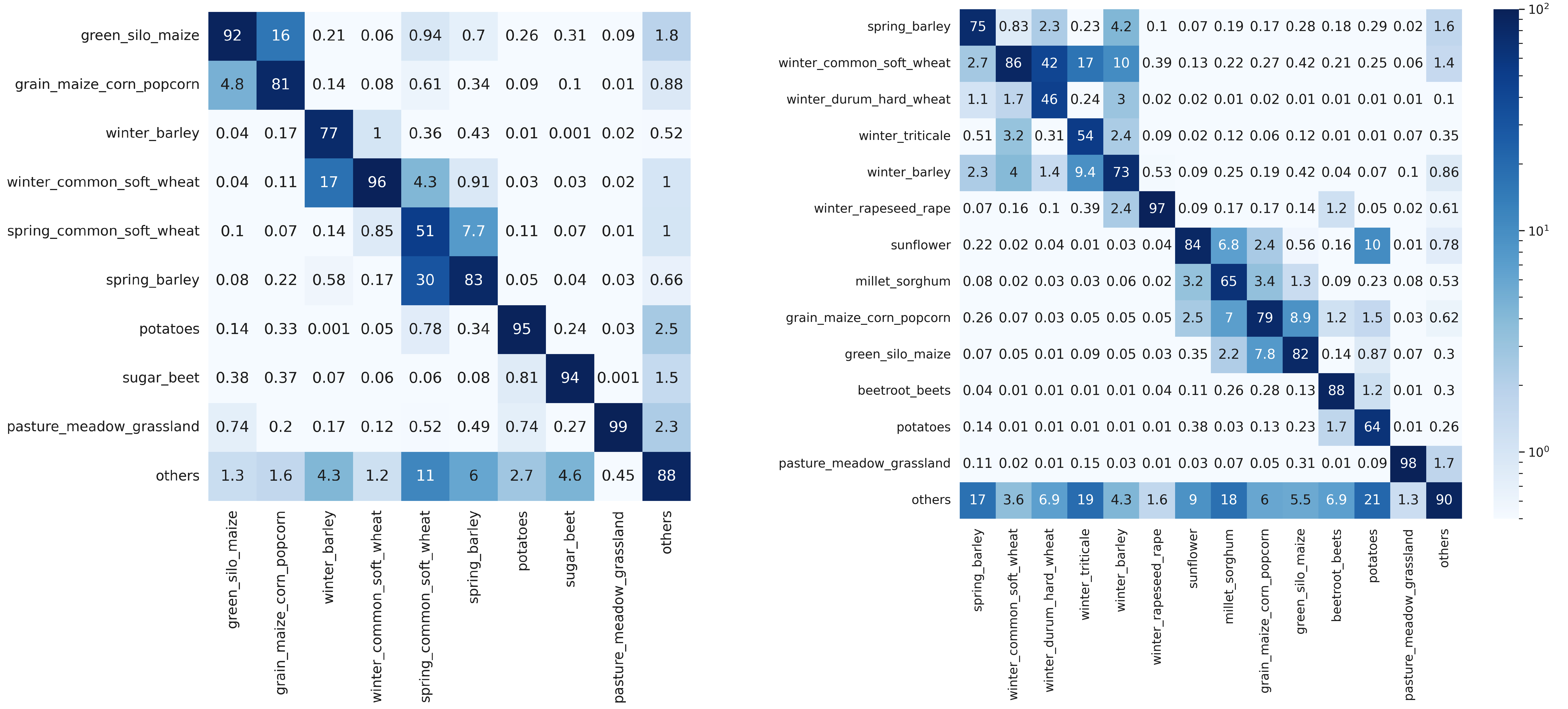} 
    \caption{Confusion matrices for the 10- and 14-class settings for Netherlands and France, respectively. True labels are in rows and predicted label are in columns, with a column-level normalization.
    }
    \label{fig:ConfusionMatrices}
\end{figure*}

The confusion matrices for both countries can be seen in \ref{fig:ConfusionMatrices}. 
Because of the highly imbalanced distribution of the labels, we classically normalised each matrix regarding the predicted values (i.e. by row) for visibility reasons. Hence, the diagonal cells represent the precision of the model on each class. We display the values in percentage, meaning they go from 0 to 100. 
These matrices allow for grouping crops based on the observed confusions. 
In general, for both countries, there are instances of misclassification between crops, indicating difficulties in distinguishing between certain crop types:
(i) between \textit{green silo maize} and \textit{grain maize corn popcorn}, 
(ii) between winter cereals (\textit{winter common soft wheat}, \textit{winter barley}, \textit{winter triticale}), 
(iii) between spring cereals (\textit{spring barley}, \textit{spring common soft wheat})
, (iv) between summer crops \textit{sunflower}, \textit{millet/sorghum}.
These confusions are anticipated as they occur with synchronous phenologies of the crops that could differs significantly from one region to the other in Europe \citep{d2020detecting,d2021parcel,meroni2021comparing}.

\subsection{Early-season models}

The use of sub-setting technique, intended for in-season classification, was found to be ineffective in improving the performance of a model solely trained on remote sensing data, as shown in \ref{fig:MF1_ModalitiesTimeSerie_NL}. However, when  the multimodal model including crop rotations was applied, it resulted in improved performances as early as May. It is worth noting that the overall performances of the multimodal model was observed to be inferior to that of a unimodal crop-only model. This was due to the model overemphasizing the RS modality as the season progressed. To address this issue, a gate mechanism could be incorporated, as proposed in \cite{Arevalo2017} and \cite{chen2017multimodal}, which selectively discards noisy modalities.
By utilizing the multimodal hierarchical configuration, the model achieved around mid-July 95\% of the end-of-season overall accuracy. This corresponds to the period when the winter crops are harvested, and the summer crops reach their peak vegetation.

In \ref{fig:F1_Crops_NL}, the $\text{HierE}_{MM}$ model data-augmented model was assessed through time. F-score are provided for the 10 classes of interest for NL, as defined in Section \ref{sec:experiments}. Overall, there was an offset of the curve rise-up. 
The shift is earlier (from mid-April to mid-June), which was consistent with other studies \citep{Russwurm2023}.

\subsection{Cross-country transfer learning results}

According to \ref{tab:FewShots_NL}, there were two notable distinctions between the models pre-trained on \ac{FR} and the ones trained from scratch. The first distinction pertained to few-shot setting, where the pre-training enables not only superior but also more consistent results in few-shot classification, reducing dependence on the specific examples used during few-shot training.
The second distinction lies in the full-data setting, where we observed that the from-scratch model performs better when validated with 141 and 24 classes (with an \ac{m-F1} score of 40.2 and 75.8, respectively, compared to 36.0 and 55.9), while the pre-trained model was better when using fewer classes. These results highlighted an intriguing behaviour, suggesting that the pre-trained neural network was more general and less prone to over-fitting on the prominent classes of the \ac{NL} dataset. We can conclude that transfer learning proves beneficial when limited labeled examples are available, while in the full-dataset training mode, it enhances the model's performance for general classes at the expense of specific classes in the dataset.

\subsection{Limitations
} 
\label{S:limitations}

The primary constraint of this study, particularly in developing countries and numerous other nations, is the requirement for digitized parcel boundary data. Another limitation is the necessity to obtain information regarding the crops cultivated in the previous season. The impact of noisy input data, such as past crop information derived from a prediction system rather than ground-truth data, remains unknown in terms of the system's response. Exploring this aspect constitutes an intriguing avenue for future research. 

For encoding the RS signal, we utilized a backbone consisting of a mean aggregation of the \ac{FOI}'s pixels, followed by the application of a temporal context window with statistical functionals. Studies proved that other methods were more efficient in terms of performances \cite{SainteFareGarnot2020}. An inherent enhancement to encode the RS signal in a more effective manner would involve employing an end-to-end approach. This approach entails learning the aggregation of RS data and integrating its representation into a neural network, similar to architectures such as \ac{CNN}-temporal or \ac{CNN}-\ac{RNN} \citep{Pelletier2019,SainteFareGarnot2019} or more advanced structures like \ac{PSE-LTAE} \citep{SainteFareGarnot2020,Quinton2021,Weilandt2023}. By adopting these powerful architectures, the encoding of RS signals can be optimized, thereby potentially improving the overall performance and accuracy of the system.

Finally, we would like to try a meta-learning method to tackle the few-shot learning more effectively, by using specific algorithm like MAML \citep{Finn2017} and MetaNorm \citep{du2020metanorm}. 
Also, a domain adaptation method like the one proposed by \cite{Capliez2023}, but spatially and not temporally, would be very useful for the few-shot setting.

\subsection{Recommendations} 
\label{sec:recomend}

Regarding future research directions, there are several avenues for further exploration. One potential direction is to integrate knowledge from the EuroCrop ontology graph inside the learning model, for example by creating multi-level embeddings of each crop. This could improve the ability of the model to capture the complex spatiotemporal variability of crops. 
It would be also possible to integrate knowledge at the loss level, in a way similar to what \cite{Turkoglu2021a} proposed. 

A more complex way to fuse the modalities together could also be explored, such as using a Gated Multimodal Unit \citep{Arevalo2017}. This could lead to better integration of the different data modalities and improved performance of the model.

It would also be valuable to investigate the results at a more regional/local level, especially for \ac{FR} with its large landmass, crop diversity, and meteorological conditions. Local hierarchical clustering and the performance of the model at the regional level could be examined to gain a deeper understanding of how the model performs in different regions.
We saw that the results for \ac{FR} were lower, possibly due to the diversity of crops and the distribution vector used. Investigating the effect of a region-specialized model, for example by fine-tuning using Adapter layers \citep{Poth2020}, could be a potential solution.
In addition, adding meteorological features and investigating their impact could be worthwhile, especially in the case of extreme events. Methods such as \cite{Tseng2021} or using learned embeddings that represent the time of the thermals \citep{Nyborg2022} could be explored. 

Another area to investigate is the potential of a specific loss function for the early season model, like the one proposed in \cite{Russwurm2023}, as per our simple data-augmentation technique. This could lead to better performance in the early season, which is an ongoing challenge for crop classification.

Other potential avenues for future work include adding more countries to the experiments, but also testing the system with different backbones, allowing ingestion of the \ac{EO} raw time series as they are. 
Furthermore, assessing the effects of different combinations of bands and sensors, encompassing various specifications such as spatial, temporal, angular, and spectral aspects, would be necessary to determine their influence on performance.
This last step would prevent reliance on man-made filters like Hampel or Whittaker and man-made features like \ac{FAPAR} and \ac{LAI} , as they contain filtered information, filtering some that may be useful for the final task \citep{Trigeorgis2016}.

Overall, these future directions could further improve the accuracy and generalization of the proposed multimodal approach for crop classification.

\section{Conclusions}

In conclusion, we proposed a multimodal hierarchical approach for crop classification that leverages crop rotation history, optical remote sensing signals, and local crop distributions. We released a large harmonized time series dataset of 7M Feature Of Interest (FOI) for a total of around 35M FOI-season. We introduced a new dataset-agnostic method relying on data and expert knowledge for aggregating crops, allowing to evaluate a classifier on a specific region in a meaningful way. Finally, we propose a data-augmentation method to boost the results in early-season setting. Our approach achieved high accuracy without in-situ data from the test season and showed promising results for cross-domain generalization through transfer learning and few-shot learning experiments. Pre-training on a dataset improves domain adaptation between countries, allowing for cross- domain and label prediction and stabilization of the learning in a few-shot setting. Our approach can contribute significantly to agriculture management and policy monitoring.

\section*{Author contributions}
V.B. and M.C. conceptualized the study. V.B., M.C. and R.D. designed the methodology: M.C. extracted the RS time-series and the original cropcodes on every Feature Of Interest, and proposed to use neural nets on rotations. V.B. proposed hierarchical multimodal models, the hierarchical aggregation, the data-augmentation, the few-shot and transfer learning experiments, extracted the features and ran the experiments. M.S. provided the EuroCrops dataset, harmonisations and support. R.D. helped to formalize all the research. V.B., M.C. and R.D. wrote the draft of the paper. All the authors analyzed the results and wrote the final paper. 

\section*{Acknowledgements}
The authors would like to thank Momtchil Iordanov for his support for visuals and Loïc Landrieu for the useful comments on the manuscript. They also would like to thank the Big Data Analytics project for their continuous support. V.B. has been funded by the grant National Center for Artificial Intelligence CENIA FB210017, Basal ANID.

\bibliography{sample.bib,JRC.bib}

\onecolumn

\clearpage

\setlength{\cftfignumwidth}{1.4cm}
\setlength{\cfttabnumwidth}{1.5cm}


\newpage

\setcounter{figure}{0}
\setcounter{table}{0}
\setcounter{page}{1}

\section*{Appendix A}
\label{AppendixA}
\captionsetup{list=no}
\renewcommand{\thetable}{Table A.\arabic{table}}
\renewcommand{\thefigure}{Fig. A.\arabic{figure}}

\begin{figure}[h]
    \centering 
    \includegraphics[width=0.8\textwidth]{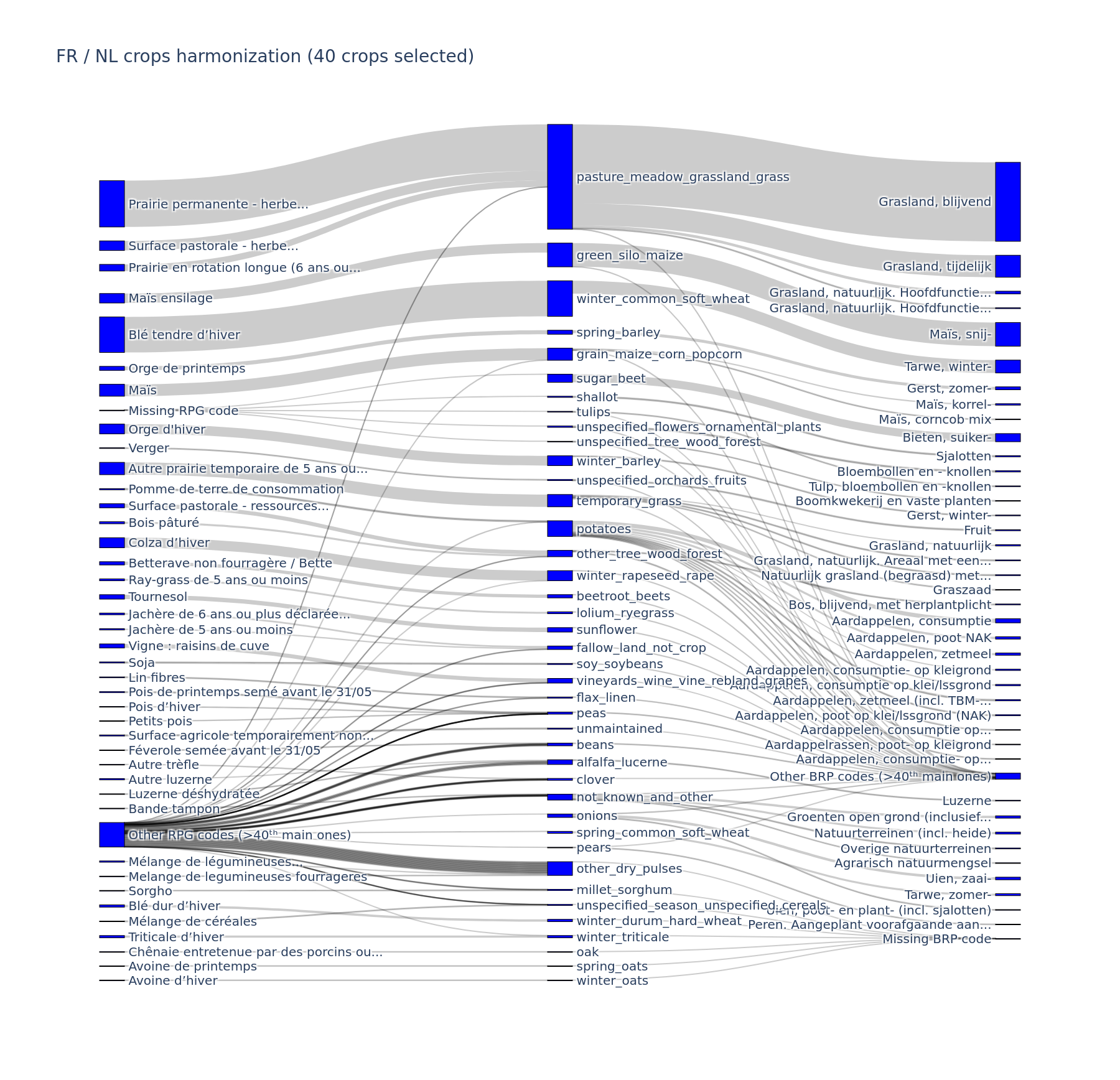} 
    \caption{Sankey diagram of the crop harmonisation, linking the French RPG (left) and the Dutch BRP (right), using HCATv2 from EuroCrops (center). The bars represent a relative share of the surface for each country. For sake of presentation (in order to fit in one page), only the 40 main crop types for each country are represented. Other crop types are grouped in "Other" classes. An interactive version of the diagram without class limitation is available on \url{https://jeodpp.jrc.ec.europa.eu/ftp/jrc-opendata/DRLL/CropDeepTrans/data/sankey_All_crops.html}.
    }
    \label{fig:Sankey}
\end{figure}

\begin{figure}[]
    \centering 
    \includegraphics[width=1\textwidth] {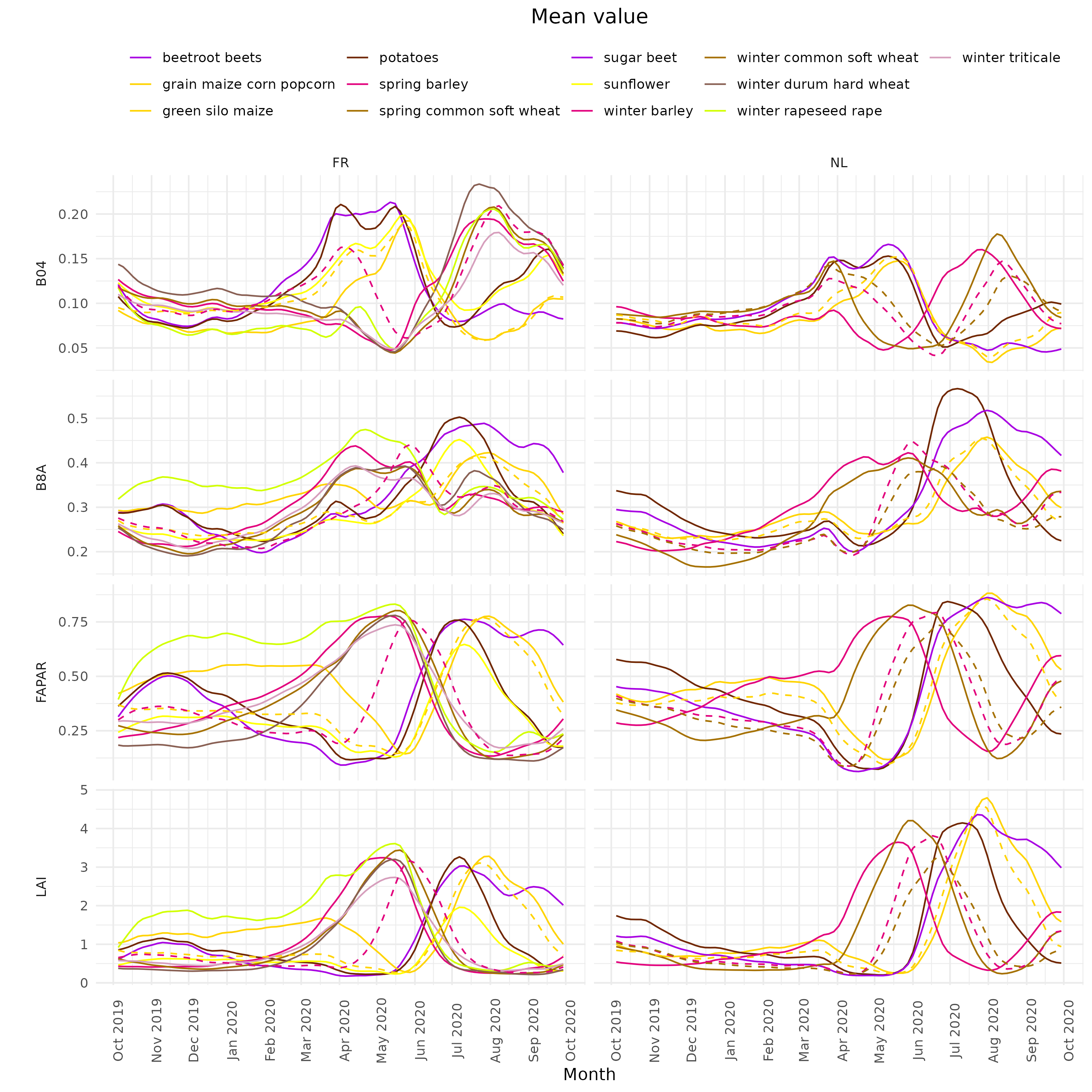} \vspace*{-.5cm}
    \caption{Mean of the EO-derived variable for each country 2019-2020. When crop have winter and spring varieties, the spring varieties are represented as dashed lines..}
    \label{fig:Mean_Crop_EO_timeSeries} 
\end{figure} 

\begin{figure}[]
    \centering 
    \includegraphics[width=1\textwidth]{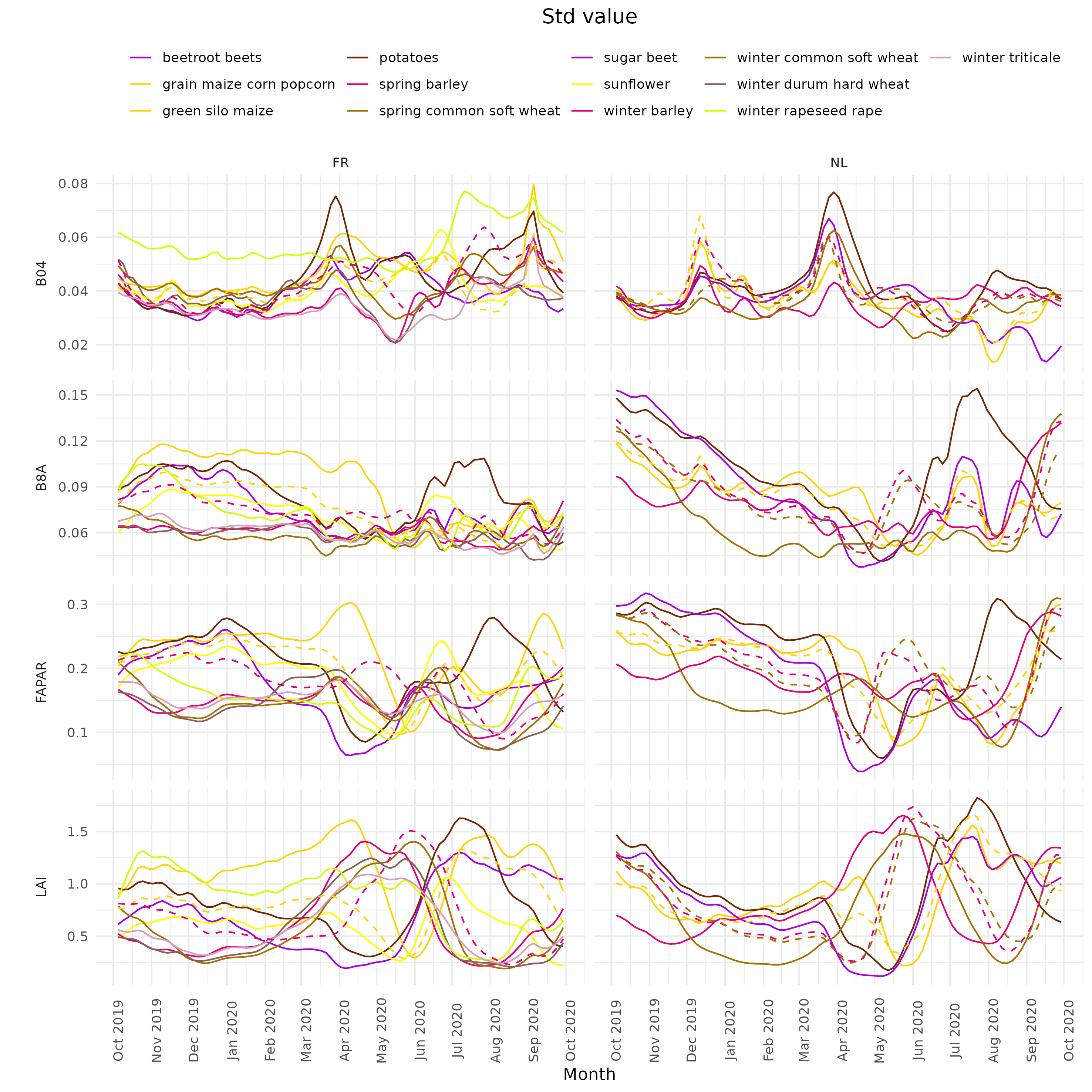} \vspace*{-.5cm}
    \caption{Standard deviation of the EO-derived variable for each country for the growing season 2019-2020. When crop have winter and spring varieties, the spring varieties are represented as dashed lines.}
    \label{fig:STD_Crop_EO_timeSeries} 
\end{figure}

\section*{Appendix B}
\label{AppendixB}

In our main paper, all the experiments were ran using a temporal split between the train, dev and test. In Table \ref{tab:Results_spacial_split} we show the results of experiments using a spatio-temporal split. It means the parcels used for testing are from different years than the year used for training, but also they are different parcels. We used 90\% of the parcels for training, and 10\% of the parcels for testing. Our results show a slight deterioration of the performances, coinciding with the ones of \cite{Weilandt2023} as we stated in the main paper. 
\begin{table}[!ht]
\centering
\resizebox{1\textwidth}{!}{
\begin{tabular}{l|c|llll|llll|llll|llll}
 \textbf{Labels} & \multirow{2}{*}{\textbf{Split}}  &  \multicolumn{4}{c}{\textbf{141/151-class}} & \multicolumn{4}{c}{\textbf{24/32-class}} & \multicolumn{4}{c}{\textbf{10/14-class}} & \multicolumn{4}{c}{\textbf{8/12-class}} \\ 
  \textbf{Dataset} & & P & R & F1 & Acc &  P & R & F1 & Acc &  P & R & F1 & Acc &  P & R & F1 & m-F1 \\  \hline \hline 
\multirow{2}{*}{NL}    & T    & 47.2 & 41.9 & 42.7 & 
93.7 & 77.2 & 75.9 & 76.0 & 94.1 & 87.0 & 82.1 & 83.8 & 95.6 & 85.7 & 78.9 & 81.4 & 91.7 \\
    & T+S  & 47.3 & 42.5 & 42.7 & 93.3 & 75.8 & 74.8 & 74.6 & 93.7 & 86.6 & 81.2 & 82.9 & 95.3 & 85.3 & 77.8 & 80.4 & 91.0 \\\hline 
\multirow{2}{*}{FR}    & T    & 44.4 & 38.7 & 39.4 & 85.4 & 72.0 & 68.6 & 69.1 & 85.6 & 79.5 & 75.9 & 77.4 & 89.1 & 77.8 & 73.2 & 75.1 & 83.5 \\
    & T+S  & 44.8 & 38.4 & 38.4 & 85.1 & 71.6 & 68.0 & 68.8 & 85.4 & 79.1 & 74.4 & 76.3 & 88.8 & 77.3 & 71.5 & 73.9 & 82.7 \\
\end{tabular}
}
\caption{Results over the Netherlands and France of the best end-of-season classification architecture on 10\% of the parcels of the dataset for the test season 2020. One split is purely temporal (T) and and the other one is temporal and spatial (T+S). The T split lines contain the results of our model on this specific 10\% subpart of the dataset. The T+S lines contain the results of a model with the same architecture trained only over the 90\% remaining. None of the models have seen timeseries from season 2020 during training. 
The metrics shown are macro Precision (P), Recall (R) and F1 score, as well as accuracy and micro-F1 score (m-F1). 
}
\label{tab:Results_spacial_split}
\end{table}

\section*{Appendix C}
\label{AppendixC}

In our main paper, all the experiments were ran using a set of features comprising FAPAR, LAI, B4 and B8A. As we stated in the main paper, the focus of this work is not on the input features sets, as our architecture is modular and the encoder we used can be changed for another one, using different features. Our work on modeling focused on showing the interest of an hierarchical architecture. 

In Table \ref{tab:Results_allbands} we show the results of experiments using more features obtained from more spectral bands, over data from \ac{NL}. 
In addition to the initial feature set, we added B2, B3, B8 given at 10m resolution, and also B11 and B12 that were interpolated at 10m. With 9 time-series instead of 4, we obtained a feature vector of size 63 instead of 28.    
Our results show an amelioration of all the models that are using more features. We also observe the same improvement between the different architectures proposed, confirming that the hierarchy helps even with a broader feature set. 

\begin{table}[!ht]
\centering
\resizebox{1\textwidth}{!}{
\centering
\begin{tabular}{l|c|llll|llll|llll|llll}
 \textbf{Labels} & \multirow{2}{*}{\textbf{Features}}  &  \multicolumn{4}{c}{\textbf{141-class}} & \multicolumn{4}{c}{\textbf{24-class}} & \multicolumn{4}{c}{\textbf{10-class}} & \multicolumn{4}{c}{\textbf{8-class}} \\ 
  \textbf{Model} & & P & R & F1 & Acc &  P & R & F1 & Acc &  P & R & F1 & Acc &  P & R & F1 & m-F1 \\  \hline \hline 
   $\text{IntraYE}_{RS}$ &  \multirow{4}{*}{4}   & 27.4 & 20.9 & 20.4 & 89.8 & 64.0 & 60.9 & 60.4 & 90.3 & 78.8 & 75.9 & 74.5 & 92.9 & 76.1 & 72.6 & 70.8 & 87.8 \\
   $\text{IntraYE}_{MM}$ &  & 55.6 & 39.7 & 43.2 & 92.8 & 76.6 & 69.8 & 72.1 & 93.1 & 83.0 & 80.5 & 80.9 & 94.7 & 80.2 & 77.9 & 78.0 & 90.0 \\
   $\text{InterYE}_{MM}$ &   & 41.1 & 33.0 & 33.6 & 92.2 & 70.8 & 70.5 & 69.9 & 92.6 & 82.2 & 79.7 & 80.4 & 94.5 & 80.2 & 76.3 & 77.5& 89.5 \\
   $\text{HierE}_{final}$ &   & 47.1 & 39.3 & 40.2 & \textbf{93.6} & 76.6 & 75.8 & 75.7 & \textbf{94.0} & 86.7 & 81.9 & 83.6 & \textbf{95.5} & 85.3 & 78.7 & 81.1 & \textbf{91.6}\\\hline 
   $\text{IntraYE}_{RS}$ &  \multirow{4}{*}{9}   & 36.0 & 27.4 & 27.4 & 92.5 & 73.3 & 68.9 & 69.6 & 92.9 & 86.5 & 82.3 & 82.2 & 95.2 & 84.7 & 79.9 & 79.6 & 91.7 \\

   $\text{IntraYE}_{MM}$ &    & 61.0 & 45.6 & 49.0 & 94.3 & 81.0 & 75.9 & 77.7 & 94.6 & 88.0 & 85.9 & 86.3 & 96.1 & 85.9 & 84.2 & 84.2 & 92.7 \\
   $\text{InterYE}_{MM}$ & & 47.1 & 35.8 & 37.9 & 94.0 & 77.2 & 74.6 & 75.5 & 94.4 & 88.6 & 84.5 & 86.0 & 96.1 & 87.3 & 82.2 & 84.0 & 92.8 \\
   $\text{HierE}_{final}$ &   & 52.5 & 46.3 & 46.7 & \textbf{95.3} & 81.5 & 81.2 & 81.1 & \textbf{95.6} & 90.6 & 87.9 & 89.0 & \textbf{96.9} & 8.96 & 85.7 & 87.3 & \textbf{94.2} \\
\end{tabular}
}
\caption{Results over the Netherlands of several end-of-season classification architectures, using different features sets. The first one is composed of B4, B8A, FAPAR and LAI, and the second one is an extension composed of first one with B2, B3, B8, B11 and B12. The metrics shown are macro Precision (P), Recall (R) and F1 score, as well as accuracy and micro-F1 score (m-F1). 
}
\label{tab:Results_allbands}
\end{table}

\section*{Appendix D}
\label{AppendixD}

The computational power in terms of training time per batch and per dataset, and GPU memory used during the training of the different models is shown in Table \ref{app:tabcomputation}. 
We used a batch size of 256. We did not use any stopping strategy, as we chose the model giving the best performance on the validation set. We used a maximum number of 25 epochs for \ac{NL} and 10 epochs for \ac{FR}. We used \texttt{float16} as precision for the weights and neurons of the neural network.

\begin{table}[!ht]
    \footnotesize
    \resizebox{1\textwidth}{!}{
    
    \centering
    \begin{tabular}{@{}l|ccc|cc|c@{}|ccc|c}
    \toprule
    \multirow{2}{*}{\textbf{Models}}    & \multirow{2}{*}{\textbf{CR}} & \multirow{2}{*}{\textbf{RS}}        & \multirow{2}{*}{\textbf{CD}} & \multicolumn{2}{c|}{\textbf{Modelisation-level}}   & \multirow{2}{*}{\textbf{Hierar.}}      & \multicolumn{3}{c|}{\textbf{Training time} (s)}    & \multirow{2}{*}{\textbf{Mem.}}  \\ 
     & & & & \textbf{Within season} & \textbf{Between seasons} & & \textbf{Batch/s} & \textbf{NL} & \textbf{FR} & \\ \midrule
    
    $\text{IntraYE}_{RS}$   & \xmark          & \cmark  & \xmark   & \cmark  & \xmark & \xmark & 49 & 3575 & 12300 & 1.03\\ 
    $\text{IntraYE}_{MM}$   & \cmark          & \cmark  & \xmark   & \cmark  & \xmark & \xmark & 47 & 3725 & 12810 & 1.03\\ \hline  
    $\text{InterYE}_{Crop}$ & \cmark          & \xmark  & \xmark   & \xmark  & \cmark & \xmark & 116 & 500 & 1720 & 1.00\\ 
    $\text{InterYE}_{RS}$   & \xmark          & \cmark  & \xmark   & \xmark  & \cmark & \xmark & 162 & 350 & 1200 & 1.00\\
    $\text{InterYE}_{MM}$   & \cmark          & \cmark  & \xmark   & \xmark  & \cmark & \xmark & 93 & 625 & 2150 & 1.00\\ \hline 
    $\text{HierE}_{RS}$     & \xmark          & \cmark  & \xmark   & \cmark  & \cmark & \cmark & 11 & 5300 & 18230  & 1.59\\
    $\text{HierE}_{MM}$     & \cmark          & \cmark  & \xmark   & \cmark  & \cmark & \cmark & 11 & 5300 & 18230 & 1.59\\
    $\text{HierE}_{final}$  & \cmark          & \cmark  & \cmark   & \cmark  & \cmark & \cmark & 11 & 5300 & 18230 & 1.60\\ \bottomrule
    \end{tabular}
    }
    \caption{Computation for each of the models: time in seconds and GPU RAM in Gigabytes} \label{app:tabcomputation}
\end{table}


\end{document}